\documentclass{article}

\usepackage{microtype}
\usepackage{graphicx}
\usepackage{booktabs} 
\usepackage{xcolor}

\usepackage{amssymb}
\usepackage{subcaption}
\usepackage{xr-hyper}
\usepackage{hyperref}

\usepackage[accepted]{icml2020}

\icmltitlerunning{Predictive Sampling with Forecasting Autoregressive Models}

\usepackage{amsmath,amsfonts,bm}

\def\eqref#1{equation~\ref{#1}}

\def\1{\bm{1}}

\def\rvh{{\mathbf{h}}}

\def\rvx{{\mathbf{x}}}

\def\rvz{{\mathbf{z}}}

\DeclareMathAlphabet{\mathsfit}{\encodingdefault}{\sfdefault}{m}{sl}
\SetMathAlphabet{\mathsfit}{bold}{\encodingdefault}{\sfdefault}{bx}{n}

\DeclareMathOperator*{\argmax}{arg\,max}

\newcommand{\std}[1]{{ \small$\pm$#1}}
\newcommand{\best}[1]{\textbf{#1}}
\usepackage{amsmath,amsfonts,bm}

\newcommand{\pFARM}{\mathrm{P}_{\mathrm{F}}}
\newcommand{\pARM}{\mathrm{P}_{\mathrm{ARM}}}
\newcommand{\forecast}{\mathrm{F}}

\newcommand{\aw}[1]{#1}

\makeatletter
\newcommand*{\addFileDependency}[1]{
  \typeout{(#1)}
  \@addtofilelist{#1}
  \IfFileExists{#1}{}{\typeout{No file #1.}}
}
\makeatother
 
\newcommand*{\myexternaldocument}[1]{%
    \externaldocument{#1}%
    \addFileDependency{#1.tex}%
    \addFileDependency{#1.aux}%
}
\myexternaldocument{supplement}

\begin{document}

\setcounter{tocdepth}{1}

\twocolumn[
\icmltitle{Predictive Sampling with Forecasting Autoregressive Models}

\begin{icmlauthorlist}
\icmlauthor{Auke Wiggers}{qual}
\icmlauthor{Emiel Hoogeboom}{uva,intern}
\end{icmlauthorlist}

\icmlaffiliation{qual}{
Qualcomm AI Research, Qualcomm Technologies Netherlands B.V.. Qualcomm AI Research is an initiative of Qualcomm Technologies, Inc.}
\icmlaffiliation{uva}{
University of Amsterdam, Netherlands.}
\icmlaffiliation{intern}{Research done while completing an internship at Qualcomm AI Research}

\icmlcorrespondingauthor{Auke Wiggers}{auke@qti.qualcomm.com}

\icmlkeywords{Machine Learning, ICML}

\vskip 0.3in
]



\printAffiliationsAndNotice{}

\begin{abstract}
Autoregressive models (ARMs) currently hold state-of-the-art performance in likelihood-based modeling of image and audio data. 
Generally, neural network based ARMs are designed to allow fast inference, but sampling from these models is impractically slow. 
In this paper, we introduce the \emph{predictive sampling} algorithm: a procedure that exploits the fast inference property of ARMs in order to speed up sampling, while keeping the model intact.
We propose two variations of predictive sampling, namely sampling with \emph{ARM fixed-point iteration} and \emph{learned forecasting modules}.
Their effectiveness is demonstrated in two settings: \textit{i)} explicit likelihood modeling on binary MNIST, SVHN and CIFAR10, and \textit{ii)} discrete latent modeling in an autoencoder trained on SVHN, CIFAR10 and Imagenet32. 
Empirically, we show considerable improvements over baselines in number of ARM inference calls and sampling speed.

\end{abstract}

\section{Introduction}
Deep generative models aim to approximate the joint distribution $\mathrm{P}(\rvx)$ of high-dimensional objects, such as images, video and audio. 
When a model of the distribution is available, it may be used for numerous applications such as anomaly detection, inpainting, super-resolution and denoising. 
However, modeling high-dimensional objects remains a notoriously challenging task.

\begin{figure}[t]
    \centering
        \includegraphics[width=.39\textwidth, trim={0 5mm 0 0}]{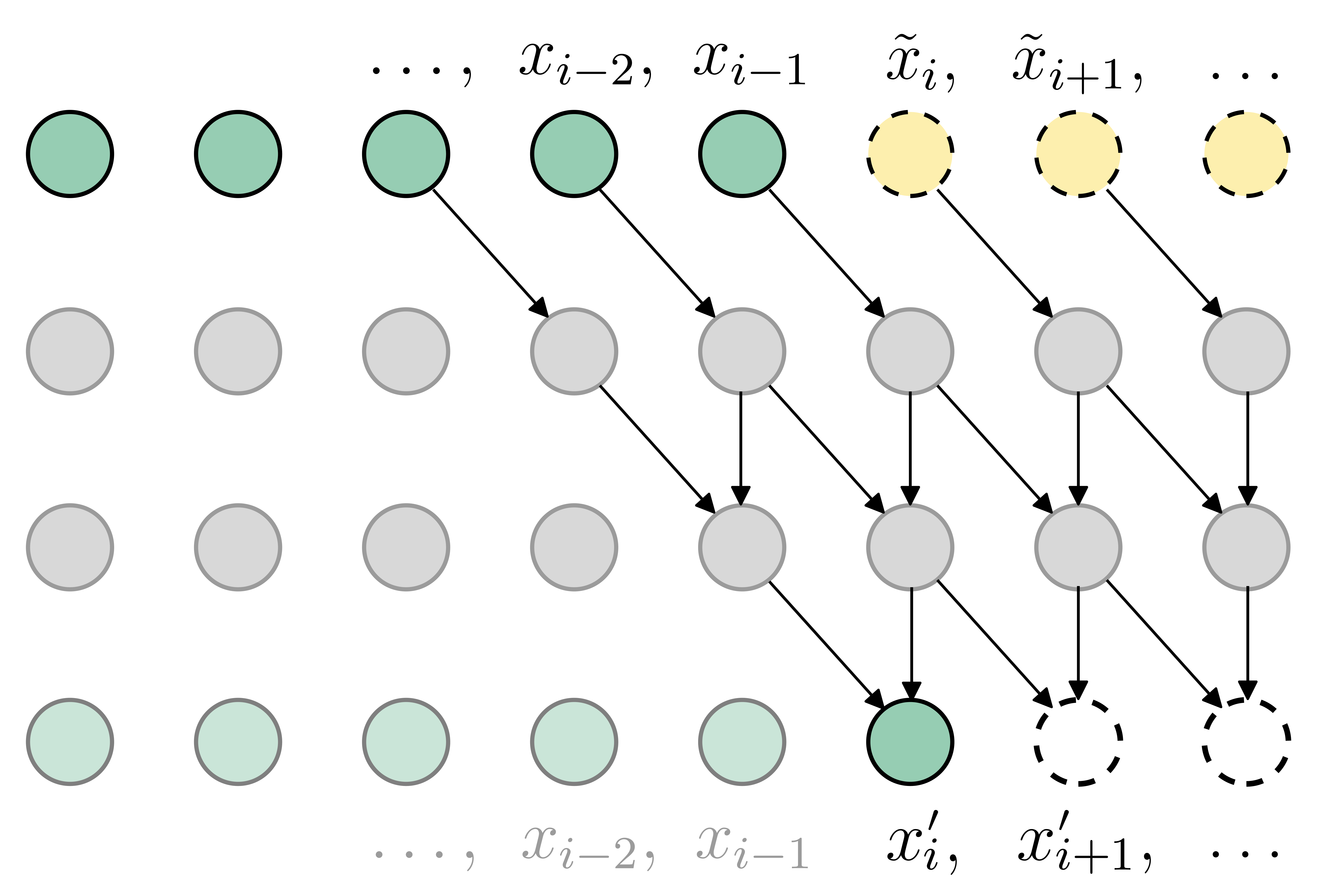}
    \caption{Overview of predictive sampling. 
             A sequence-so-far of ARM samples $x_{0},\ldots,x_{i-1}$ is extended with forecasts $\tilde{x}_i, \tilde{x}_{i+1}, 
    \ldots$ and given as input to the ARM. 
    As the ARM has strict triangular dependence, its first output $x'_i$ is valid: the conditioning consists only of ARM samples.
    If the forecast $\tilde{x}_i$ is equal to $x'_i$, the next output $x'_{i+1}$ is also valid. 
    }
    \label{fig:first_image}
    \vspace{-5mm}
\end{figure}

A powerful class of distribution models called deep autoregressive models (ARMs) \cite{bengio2000taking,larochelle2011neural} decomposes the high-dimensional joint distribution into single-dimensional conditional distributions, using the chain rule from probability theory. 
Neural network based ARMs currently hold state-of-the-art likelihood in image and audio domains \citep{oord2016wavenet, oord2016pixel, salimans2017pixelcnn, chen2018pixelsnail, menick2018generating, child2019generatingsparse}.

A major limitation of ARMs is that the autoregressive computation can be parallelized only in a single direction: either \emph{evaluation} or \emph{sampling}. 
Generally, these models are trained using likelihood evaluation and 
\aw{empirical computational cost of training is substantially higher than that of sampling}.
As such, ARMs are designed to allow for fast evaluation, but sampling from these models is prohibitively slow. 
In the literature, there are methods that try to accelerate sampling by breaking autoregressive structure \emph{a priori}, but consequently suffer from a decrease in likelihood performance \citep{reed2017parallel}. 
Another approach approximates an autoregressive density model via distillation \cite{oord2018parallelwavenet}, but this method provides no guarantees that samples from the distilled model originate from the original model distribution.

\begin{figure*}[t]
    \centering
    \begin{subfigure}[trim={0 3mm 0 0}]{0.47\textwidth}
        \centering
        \includegraphics[width=0.9\textwidth]{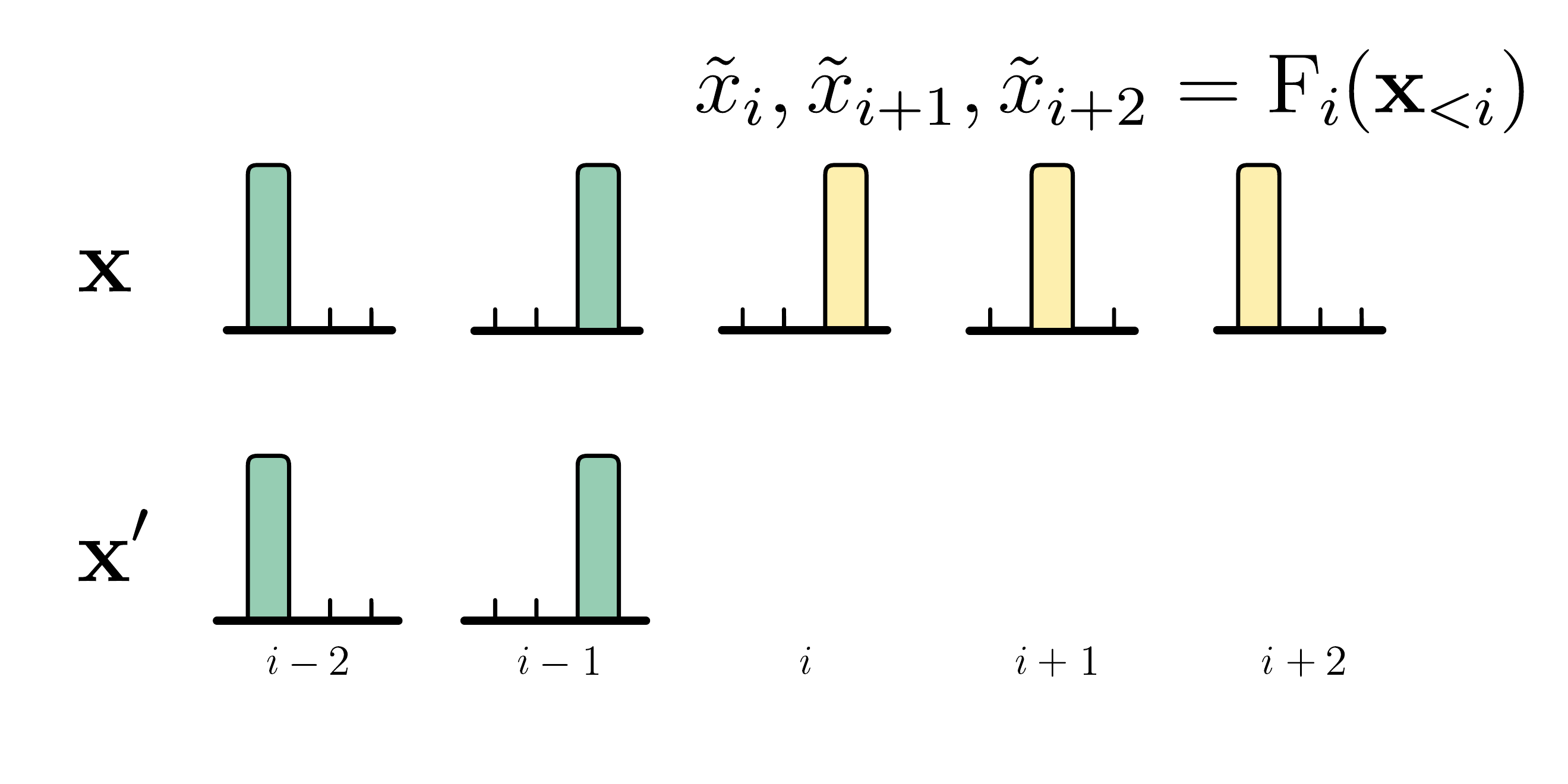}
        \vspace{-2mm}
        \caption{Given is a sequence-so-far $\rvx_{<i}$ for a position $i$, here one-hot encoded. A future sequence $\tilde{x}_i, \tilde{x}_{i+1}, \ldots$ is predicted by forecasting function $\forecast_i$ and is appended to $\rvx$.}
    \end{subfigure}
    \hspace{.5cm}
    \begin{subfigure}[]{0.47\textwidth}
        \centering
        \includegraphics[width=0.9\textwidth]{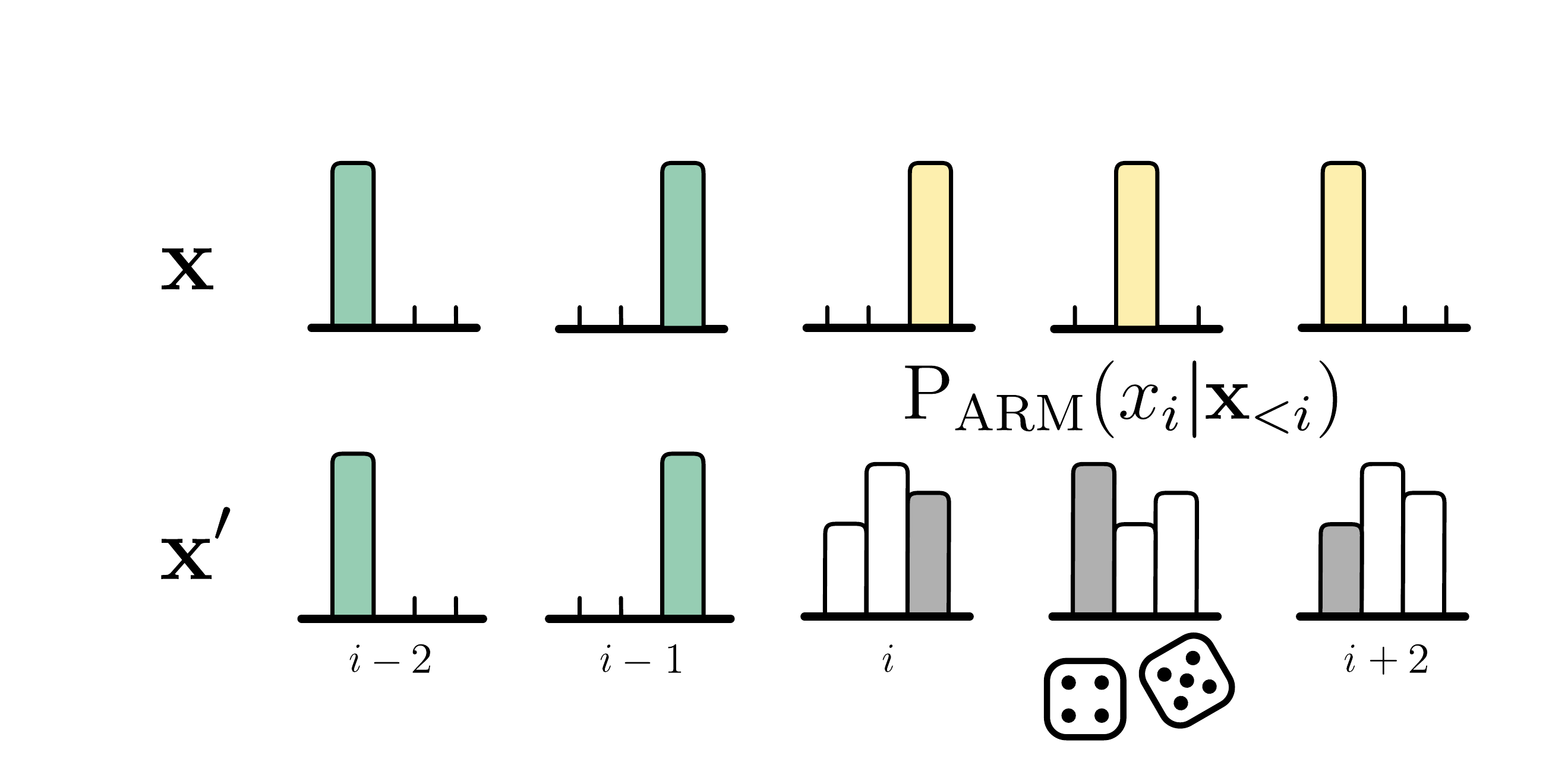}
        \vspace{-2mm}
        \caption{An inference pass of the model $\pARM$ conditioned on $\rvx$ gives the distribution parameters for every time step, which allows sampling $x'_i, x'_{i+1}, \ldots$, the outputs of the ARM.}
    \end{subfigure}
    \begin{subfigure}[trim={0 3mm 0 0}]{0.47\textwidth}
        \centering
        \includegraphics[width=0.9\textwidth]{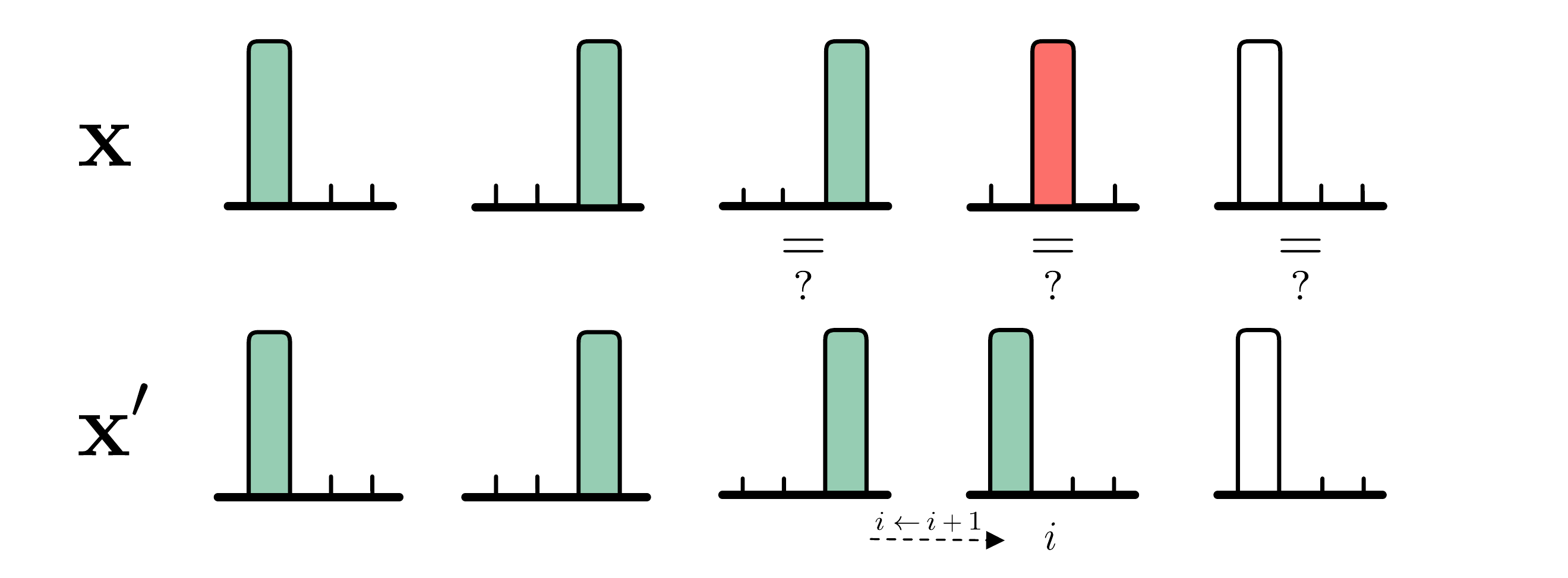}
        \vspace{-2mm}
        \caption{For each forecast in $x_{i}$ that equals the sampled value $x'_{i}$, the position $i$ is incremented, without any additional inference calls. This process repeats until $x_{i} \not= x'_{i}$. If at some point $i=d$ the algorithm returns. 
        }
    \end{subfigure}
    \hspace{.5cm}
    \begin{subfigure}[]{0.47\textwidth}
        \centering
        \includegraphics[width=0.9\textwidth]{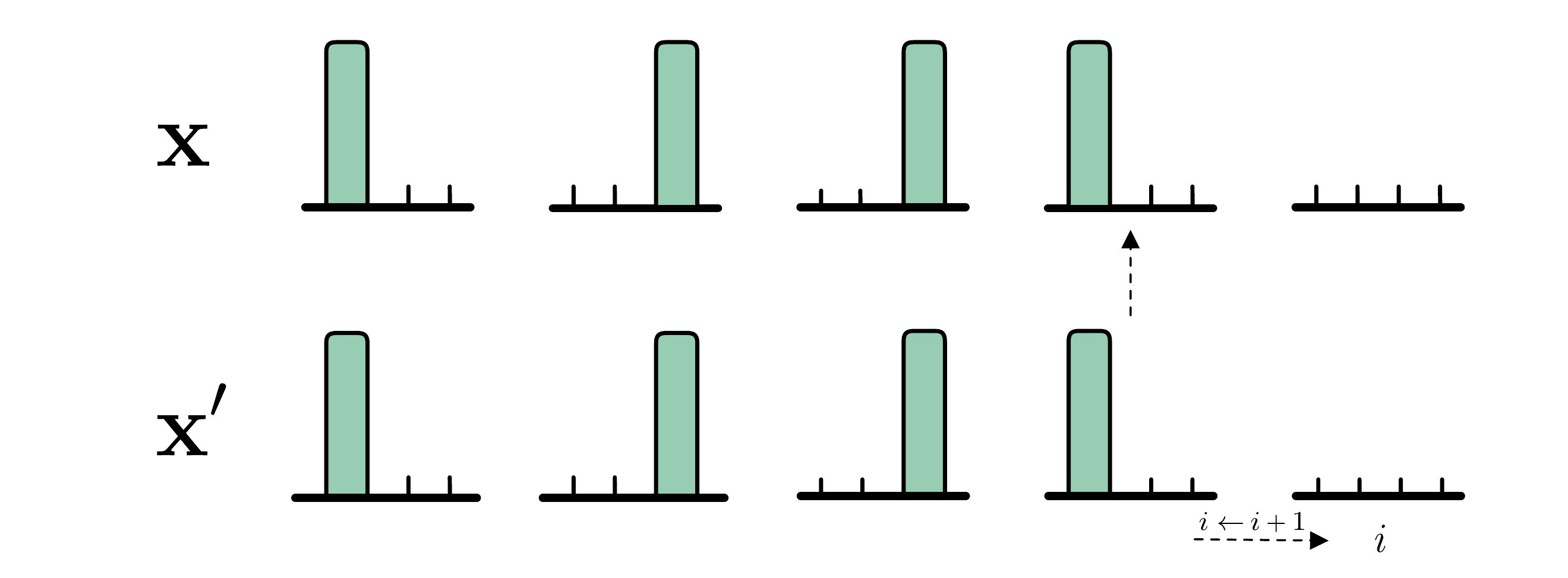}
        \vspace{-2mm}
        \caption{Although $x_{i} \not= x'_{i}$, the output $x'_i$ is a valid sample, as its distribution depends on $\rvx_{<i}$, which is equal to $\rvx'_{<i}$. 
        The value is written to the forecast vector, $i$ is incremented by one, and the process repeats from (a).}
    \end{subfigure}
    \caption{One iteration of predictive sampling with forecasting autoregressive models.}
    \label{fig:overview}
\end{figure*}

This paper proposes a new algorithm termed predictive sampling, which \textit{1)} accelerates discrete ARM sampling, \textit{2)} keeps autoregressive structure intact, and \textit{3)} samples from the \emph{true} model distribution. 
Predictive sampling forecasts which values are likely to be sampled, and uses the parallel inference property of the ARM to reduce the total number of required ARM forward passes.
To forecast future values, we introduce two methods: \emph{ARM fixed-point iteration} and \textit{learned forecasting}.
These methods rely on two insights: \textit{i}) the ARM sampling procedure can be reparametrized into a deterministic function and independent noise, and \textit{ii}) activations of the penultimate layer of the ARM can be utilized for computationally efficient forecasting.
We demonstrate a considerable reduction in the number of forward passes, and consequently, sampling time, on binary MNIST, SVHN and CIFAR10. 
Additionally, we show on the SVHN, CIFAR10 and Imagenet32 datasets that predictive sampling can be used to speed up ancestral sampling from a discrete latent autoencoder, when an ARM is used to model the latent space.
For a visual overview of the method, see Figure \ref{fig:first_image}. 

\section{Methodology}

Consider a variable $\rvx \in \mathcal{X}$, where $\mathcal{X}$ is a discrete space, for example $\mathcal{X} = \{0, 1, \ldots, 255\}^d$ for 8 bit images, where $d$ is the dimensionality of the data. An autoregressive model views $\rvx$ as a sequence of 1-dimensional variables $(x_i)_{i=1}^{d}$, which suggests the following universal probability model \cite{bengio2000taking,larochelle2011neural}:
\begin{align}
    \pARM(\rvx) &= \prod_{i=1}^{d} \pARM(x_i | \rvx_{<i}),
\end{align}
Universality follows from the chain rule of probability theory, and $\rvx_{<i}$ denotes the values $(x_j)_{j=1}^{i-1}$. 
Samples from the model $\rvx \sim \pARM$ are typically obtained using ancestral sampling:
\begin{equation}
    x_i \sim \pARM( x_i | \rvx_{<i}) \quad \text{for} \quad i = 1,\ldots,d
\end{equation} 
In practice, ARMs can be implemented efficiently using deep neural networks. 
Let $f$ be a strictly autoregressive function such that when $\rvh = f(\rvx)$, the representation $\rvh_i$ depends only on input values 
$\rvx_{<i}$ 
. 
The parameters $\boldsymbol{\theta}_i$ for the distribution over $x_i$ are then an autoregressive function of the representation such that $\boldsymbol{\theta}_i$ depends on $\rvh_{\leq i}$. 
Using this formulation it is possible to parallellize ARM inference, \textit{i.e.}, to obtain a log-likelihood for every variable in parallel. 
However, in this setting, na\"{i}ve sampling from an ARM requires $d$ forward calls that cannot be parallelized.

\subsection{Predictive Sampling}
Consider the na\"{i}ve sampling approach.
First, it computes an intermediate ARM representation $h_1 = f_1(\varnothing)$ and distribution parameters $\theta_1(h_1)$, then samples the first value $x_1 \sim \mathrm{P}(x_1 | \theta_1)$.
Only then can the next representation $h_2 = f_2(x_1)$ be computed. 

In the setting of \emph{predictive sampling}, suppose now that we can obtain a \emph{forecast} $\tilde{x}_1$, which is equal to $x_1$ with high probability. 
In this case $[ h_1, h'_2 ] = f(\tilde{x}_1)$ can be computed in parallel.
\aw{We say that} $h'_2$ is \emph{valid}, \textit{i.e.}, it is equal to $h_2$, if $\tilde{x}_1$ equals $x_1$.
We proceed as before and sample $x_1 \sim \mathrm{P}(x_1 | \theta_1(h_1))$. 
If the sampled $x_1$ is indeed equal to the forecast $\tilde{x}_1$, the representation $h'_2$ is valid, and we can immediately sample $x_2 \sim \mathrm{P}(x_2 | \theta_2(h_1, h_2))$ without additional calls to $f$.
More generally, for a sequence of $n$ correct forecasts, $n$ re-computations of $f$ are saved. 
A general description of these steps can be found in Algorithm~\ref{alg:forecasting}, where $\circ$ denotes vector concatenation and $\mathrm{F}_i$ denotes a function that outputs forecasts starting at location $i$.
\aw{Note that the ARM returns probability distributions 
$\mathrm{P}_{\mathrm{ARM}}(\,\cdot\, | \rvx)$
for all locations simultaneously, even if the input partially consists of placeholder values}. 
A corresponding visualization is shown in Figure~\ref{fig:overview}.

\begin{algorithm}[t!]
   \caption{Predictive Sampling}
   \label{alg:forecasting}
\begin{algorithmic}
\renewcommand{\COMMENT}[1]{\textcolor{blue}{\,\,\,\,\,\small#1}}
  \newcommand{\xbuffer}[0]{\rvx}
   \STATE {\bfseries Input:} \text{ARM} $\mathrm{P}_{\mathrm{ARM}}$, \text{forecasting function} $\forecast$
   \STATE {\bfseries Output:} $\xbuffer$
   \STATE let $i\gets0$, $\rvx \gets \mathbf{0}$ 
   \WHILE{$i < d$}
     \STATE $\tilde{x}_{i}, \tilde{x}_{i+1}, \ldots \gets \forecast_i (\xbuffer)$ \COMMENT{//~Forecast}
     \STATE $\xbuffer \gets \xbuffer_{<i} \circ [\tilde{x}_{i}, \tilde{x}_{i+1}, \ldots]$
     \STATE $x'_i, x'_{i+1}, \ldots \sim \mathrm{P}_{\mathrm{ARM}}(\,\cdot\, | \xbuffer)$
     \STATE \textcolor{blue}{\small //~While forecast $\tilde{x}_i$ is correct, output $x'_{i+1}$ is valid}
     \WHILE{$\tilde{x}_{i} = x'_{i}$ \AND $i < d$ }  
         \STATE $i \gets i + 1$   
     \ENDWHILE
     \IF{$i < d$}  
         \STATE $\xbuffer[i] \gets x'_{i}$ \COMMENT{//~Overwrite the input vector}
         \STATE $i \gets i + 1$
     \ENDIF
\ENDWHILE
\end{algorithmic}
\end{algorithm}

\subsection{Forecasting}
The \emph{forecasting function} $\forecast_i$ can be formalized as follows.
Specifically, consider the vector $\rvx$ in Algorithm \ref{alg:forecasting}, which contains valid samples until position $i$, \textit{i.e.}, the variables $\rvx_{<i}$ are valid samples from the ARM. 
Let $\tilde{x}_{i}$ denote a forecast for the variable $x_{i}$. 
A forecasting function $\forecast_i$ aims to infer the most likely future sequence starting from position $i$ given all available information thus far:
\begin{equation}
   \tilde{x}_{i}, \tilde{x}_{i+1}, \ldots = \forecast_i(\rvx).
\end{equation}
Using this notion, we can define the predictive sampling algorithm, given in Algorithm~\ref{alg:forecasting}.
It uses the forecasting function $\forecast_i$ to compute a future sequence $[\tilde{x}_{i}, \tilde{x}_{i+1}, \ldots]$, and combines known values $\rvx_{<i}$ with this sequence to form the ARM input.
Utilizing this input the ARM can compute an output sequence for future time steps $x'_i, x'_{i+1}, \ldots$ in parallel.
The ARM output $x'_{i}$ is \emph{valid}: it is a sample from the true model distribution, as the conditioning $\rvx_{<i}$ does not contain forecasts.
For each consecutive step $t$ where forecast $\tilde{x}_{i+t}$ is equal to the ARM output $x'_{i+t}$, the subsequent output $x'_{i+t+1}$ is valid as well. 
When the forecasting function does not agree with the ARM, we write the last valid output to the input vector $\rvx$, and proceed to the next iteration of predictive sampling.
This process is repeated until all $d$ variables have been sampled.

\paragraph{Isolating stochasticity via reparametrization}
\label{sec:method:reparametrizing_arms}
The sampling step introduces unpredictability for each dimension, which may fundamentally limit the number of subsequent variables that can be predicted correctly. 
For example, even if every forecasted variable has a $60\%$ chance of being correct, the expected length of a correct sequence will only be $\sum_{k=1}^\infty 0.6^k = 1.5$. 

To solve this issue, we reparametrize the sampling procedure from the ARM using a deterministic function $g$ and a stochastic noise variable $\boldsymbol{\epsilon}$. 
A sample $x_i \sim \pARM(\,\cdot\, | \rvx_{<i})$ can equivalently be computed using the deterministic function $g_i$ conditioned on random noise $\boldsymbol{\epsilon}$:
\begin{equation}
    x_i = g_i(\rvx_{<i}, \boldsymbol{\epsilon}) \quad \text{ where } \quad \boldsymbol{\epsilon} \sim \mathrm{p}(\boldsymbol{\epsilon}).
    \label{eq:arm_reparametrization}
\end{equation}
Such a reparametrization always exists for discrete distributions.
As a consequence the sampling procedure from the ARM has become \emph{deterministic} \aw{given noise $\boldsymbol{\epsilon}$}. 
This is an important insight, because the reparametrization resolves the aforementioned fundamental limit of predicting a stochastic sequence. 
For instance, consider an ARM that models categorical distributions over $\rvx$ using log-probabilities $\boldsymbol{\mu}$. 
One method to reparametrize categorical distributions is the Gumbel-Max trick \cite{gumbel1954statistical}, which has recently become popular in machine learning \cite{maddison2014sampling,maddison2017concrete,jang2017categorical,kool2019stochastic}. 
By sampling standard Gumbel noise $\boldsymbol{\epsilon} \sim G^{d \times K}$ the categorical sample $x_i \sim \pARM(\,\cdot\, | \rvx_{<i})$ can be computed using:
\begin{equation}
    x_i = \argmax_c \left( \vphantom{\sum} \mu_{i,c}(\rvx_{<i})  + \epsilon_{i,c} \right),
\end{equation}
\aw{
  where $c \in \{1, \ldots, K\}$ represents a category and 
  $\mu_{i,c} = \log \pARM(x_i = c | \rvx_{<i})$
  is its log probability for location $i$. 
}

\aw{
Although we focus on the discrete setting here, reparametrization noise can be obtained for many common continuous probability distributions \cite{ruiz2016generalized}.
Note that in the continuous setting, verifying that the forecast is equal to the ARM output will depend on numerical precision, and a margin of error must be used.
}

\paragraph{Shared Representation}
In theory, the future sequence can be predicted perfectly using $\boldsymbol{\epsilon}$ and $\rvx_{<i}$, as it turns the ARM into a deterministic function $g$. 
In practice however, the amount of computation that is required to predict the future sequence perfectly may exceed the computational cost of the ARM. 

\aw{
Let $m$ be the new number of iterations that predictive sampling requires to converge. 
It is desirable to design $\forecast$ such that the added computational cost of $\forecast$ is lower than the computational cost that was saved by the reduced number of ARM calls. 
Specifically, this requires that $m \cdot \mathrm{cost}(\forecast) \leq (d-m) \cdot \mathrm{cost}(\mathrm{ARM})$.
}

\aw{
To keep the computational cost of $\forecast$ low,
} 
the ARM representation $\rvh = f(\rvx)$ from the previous iteration of predictive sampling is shared with the forecasting function:
\begin{equation}
   \tilde{x}_{i}, \tilde{x}_{i+1}, \ldots = \forecast_i(\rvx, \rvh, \boldsymbol{\epsilon}).
   \label{eq:forecasting_sharing}
\end{equation}
When forecasts starting from location $i$ are required, the variables $\rvx_{<i}$ are already valid model outputs. 
In the previous iteration of predictive sampling, input $\rvx_{<i-1}$ was valid, and therefore the representation $\rvh_{<i}$ is valid as well.
Although it is possible to obtain an unconditional forecast for $i=0$, we use a zero vector as initial forecast.

\aw{
  Conditioning on $\rvh$ should in theory have no effect on performance, as the data processing inequality states that no post-processing function can increase the information content.
  However, in practice $\rvh$ is a convenient representation that summarizes the input at no extra computational cost.
}

\subsection{ARM Fixed-Point Iteration}
\aw{The first forecasting method we introduce is ARM Fixed-Point Iteration (FPI)}, which utilizes the ARM itself as forecasting function. 
Specifically, a forecast $\tilde{x}_{i+t}$ at step $t$ is obtained using the ARM reparametrization $g$, where noise $\boldsymbol{\epsilon}$ is isolated:
\begin{equation}
  \tilde{x}_{i+t} = g_{i+t}(\rvx_{<{i+t}}, \boldsymbol{\epsilon}) \,\,\text{for}\,\, t = 0, 1, \ldots
  \label{eq:arm_forecasting}
\end{equation}
Note that $\rvx_{<{i+t}}$ is a concatenation of the valid samples thus far $\rvx_{<i}$ and the forecasts $\tilde{\rvx}_{i:i+t-1}$ from the \emph{previous} iteration of predictive sampling (as in Algorithm~\ref{alg:forecasting}).
In other words, current forecasts are obtained using ARM inputs that may turn out to be invalid.
Nevertheless, the method is compelling because it is computationally \mbox{inexpensive} and requires no additional learned components: it simply reuses the ARM output.

Interestingly, the combination of forecasting with Equation \ref{eq:arm_forecasting} and Algorithm \ref{alg:forecasting} is equivalent to a reformulation as a fixed-point iteration using the function $g$ \aw{defined over $\rvx$}:
\begin{equation}
    \rvx^{(n+1)} = g(\rvx^{(n)}, \boldsymbol{\epsilon}),
\end{equation}
where $n$ denotes the iteration number of predictive sampling.
We show this reformulation in Algorithm~\ref{alg:fixedpoint}. 
This equivalence follows because ARM outputs $x'_{i}$ are fixed if their conditioning consists of samples $\rvx_{<i}$ that are valid, \textit{i.e.}, for $\rvx' = g(\rvx, \boldsymbol{\epsilon})$ the outputs $\rvx'_{<i}$ equal the inputs $\rvx_{<i}$. The future outputs are automatically used as forecasts. 
The algorithm is guaranteed to converge in $d$ steps because the system has strictly triangular dependence, and may converge much faster if variables do not depend strongly on adjacent previous variables.

\begin{algorithm}[tb]
   \caption{ARM Fixed-Point iteration}
   \label{alg:fixedpoint}
\begin{algorithmic}
  \newcommand{\xbuffer}[0]{\rvx}
   \STATE {\bfseries Input:} $g$, $\boldsymbol{\epsilon}$
   \STATE {\bfseries Output:} $\xbuffer$
   \STATE let $\xbuffer^{(0)} = \mathbf{0}$, $n = 0$
  \REPEAT
     \STATE $\xbuffer^{(n+1)} = g(\xbuffer^{(n)}, \boldsymbol{\epsilon})$
     \STATE $n = n + 1$
  \UNTIL{$\xbuffer^{(n)} = \xbuffer^{(n-1)}$}
\end{algorithmic}
\end{algorithm}

\subsection{Learned forecasting}
ARM fixed-point iteration makes use of the fact that the ARM outputs distributions $\pARM(x_i| \rvx_{<i})$ for every location $i\in\{ 1, \ldots, d\}$. 
However, many output distributions are conditioned on forecasts $\tilde{\rvx}_{i:i+t-1}$ from the previous iteration of predictive sampling, and these may turn out to be incorrect.
For example, if in the first iteration of 
the algorithm
we find that forecast $\tilde{x}_1$ does not match $x_1$, the procedure will still use the sampled ${x}'_2 \sim \pARM(x_2 | \tilde{h}_1( \tilde{x}_1 ))$ as input in the second iteration.
In turn, this may result in an incorrect forecast $\tilde{x}_3$.
In the worst case, this leads to cascading errors, and $d$ ARM inference calls are required.

To address this problem,
we introduce \emph{learned forecasting}, an addition to ARM fixed-point iteration.
We construct \emph{forecasting modules}: small neural networks that are trained to match the distribution $\pARM(x_i| \rvx_{<i})$. 
\aw{These networks may only utilize information that will be available during sampling.}
For that reason, they are only conditioned on the available \emph{valid} information, $\rvx_{<i}, \rvh_{<i}$ and $\boldsymbol{\epsilon}$. 

\aw{In particular, a forecasting module at timestep $i+t$ outputs a distribution $\pFARM^{(t)}(\tilde{x}_{i+t} | \rvx_{<i})$ that will be trained to match the ARM distribution at that location $\pARM(x_{i+t}| \rvx_{<{i+t}})$.
The important difference is that $\pFARM$ is conditioned only on $\rvx_{<i}$ and $\rvh_{<i}$, whereas the ARM output for that location is based on $\rvx_{<{i+t}}$.
We minimize the KL divergence between corresponding distributions $\pARM$ and $\pFARM$:
}
\begin{equation}
  \sum_i \mathrm{KL} \left[ 
      \vphantom{\sum} \pARM(x_{i+t} | \rvx_{<{i+t}}) ~||~ \pFARM^{(t)}(x_{i+t} | \rvx_{<i}) 
   \right] ,
   \label{eq:kl_forecasting}
\end{equation}
with respect to the forecasting module $\pFARM^{(t)}$ for each future step $t$. 
The gradient path to the model $\pARM$ is detached in this divergence.

After training, forecasts can be obtained via the forecasting distributions and reparametrization noise. 
For example, when $\pARM$ and $\pFARM$ are categorical distributions:
\begin{align}
\begin{split}
    \tilde{x}_{i+t}
    &= \argmax_c \left( \vphantom{\sum} \tilde{\mu}_{i,t,c}( \rvx_{<i} ) + \epsilon_{i,c} \right),
\end{split}
\label{eq:forecast_arm_categorical}
\end{align}
where $\tilde{\mu}_{i,t,c}(\rvx_{<i}) = \log ~ \pFARM(x_{i+t}=c | \rvx_{<i} )$ is the log-probability that $x_{i+t} = c$ according to the forecasting distribution. 
In practice, a sequence of forecasts is obtained by concatenating forecasting modules $\forecast_{i,t}$, where $t = 0, \ldots, T-1$ and $T$ is the window in which we forecast future values.

We find that explicitly conditioning on $\rvx_{<i}$ in combination with $\rvh_{<i}$ does not result in a noticeable effect on performance for the forecasting module capacity we consider. 
Instead it suffices to solely condition on $\rvh_{<i}$.
The representation $\rvh$ is shared and trained jointly for both the ARM and the forecasting modules, but the forecasting objective is down-weighed with a factor of $0.01$ so that the final log-likelihood performance is not affected. 
While it is possible to train forecasting modules on samples from the model distribution, we only train on samples from the data distribution as the sampling process is relatively slow.

\section{Related work}

Neural network based likelihood methods 
in generative modelling can broadly be divided into VAEs \cite{kingma2014autoencoding, rezende2014stochastic}, Flow based models \cite{dinh2017realnvp} and autoregressive models \cite{bengio2000taking,larochelle2011neural}. 
VAEs and Flows are attractive when fast sampling is important, as they can be constructed without autoregressive components that need inverses. 
However, in terms of likelihood performance, ARMs currently outperform VAEs and Flows and hold state-of-the-art in image and audio domains \cite{oord2016pixel,oord2016wavenet,salimans2017pixelcnn,chen2018pixelsnail, child2019generatingsparse}.

One of the earliest neural network architectures for autoregressive probability estimation of image data is NADE \cite{larochelle2011neural}. 
This model employs a causal structure, \textit{i.e.}, nodes of the network are connected in such a way that layer output $h_{i}$ only depends on a set of inputs $x_{<i}$.
Numerous follow up works by \citet{germain2015made,oord2016pixel,akoury2017spatial,salimans2017pixelcnn,menick2018generating,sadeghi2019pixelvae} improve on this idea, and increase likelihood performance by refining training objectives and improving network architectures. 

There are various approaches that aim to capture the performance of the ARM while keeping sampling time low.
The autoregressive dependencies can be broken between some of the dimensions, which allows some parts of the sampling to run in parallel, but comes at the cost of decreased likelihood performance \cite{reed2017parallel}. 
It is possible to train a student network using distillation \cite{oord2018parallelwavenet}, but in this case samples from the student network will not come from the (teacher) model distribution.

An alternative method that does preserve the model structure relies on caching of layer activations to avoid duplicate computation \cite{ramachandran2017fastgeneration}.
\aw{
  To apply this algorithm, the caching strategy must be specified in accordance with the ARM architecture.
  In addition, activations of the network must be stored during sampling, resulting in larger memory overhead.
  In contrast, our method does not require knowledge of model-specific details and does not require extra memory.
}


\aw{Finally, a method that predicts what a language model will output in order to save runtime has been proposed in \cite{stern2018blockwise}.
A key difference between this method and ours is that we sample from the model distribution instead of decoding greedily via argmax.}
\section{Experiments}

\begin{table*}
    \centering
    \caption{Performance of predictive sampling for ARMs trained on explicit likelihood modeling tasks, in terms of percentage of forward passes with respect to the original sampling procedure, and total time to sample. 
    All reported times are based on own implementation.
    Reported means and (Bessel-corrected) standard deviations are based on sampling of 10 batches with random seeds $\{ 0, \ldots, 9 \}$.
    }
    \label{tab:performance_explicit}
    \scalebox{0.875}{
    \begin{tabular}{l p{3.5cm} r r r r r r r }
    \toprule
      &                        & \multicolumn{3}{c}{Batch size 1} & \multicolumn{3}{c}{Batch size 32} \\
      &                        & ARM calls     & Time (s)         & Speedup & ARM calls       & Time (s)  & Speedup \\
      \midrule                                                                              
    MNIST (1 bit) 
      & Baseline                 & 100.0\%\std{0.0} & 16.6\std{0.1} & $1.0\times$ & 100.0\%\std{0.0}  & 24.1\std{~~0.4} & $1.0\times$ \\
      & Forecast zeros           & 14.5\%\std{5.0} & 2.4\std{0.8} & $6.9\times$ & 25.0\%\std{0.1}  &  7.6\std{~~1.0} & $3.3\times$ \\  
      & Predict last             & 7.8\%\std{1.5} & 1.5\std{0.4} & $11.1\times$ & 10.0\%\std{0.6} & 3.8\std{~~0.3} & $6.3\times$ \\
      & Fixed-point iteration    & \best{3.3}\%\std{0.9} & \best{0.6}\std{0.1} & \best{27.6}$\times$ & 5.2\%\std{0.4} & \textbf{2.8}\std{~~0.2} & \best{8.6}$\times$ \\
      & + Forecasting ($T=20$)   & \best{3.3}\%\std{0.6} & 0.7\std{0.2} & $23.7\times$ & \best{4.3}\%\std{0.3} & \textbf{2.8}\std{~~0.5} & \best{8.6}$\times$  \\
      \midrule                                                                                                    
    SVHN (8 bit)                
      & Baseline                & 100.0\%\std{0.0} & 145.7\std{0.8} & $1.0\times$ & 100.0\%\std{0.0} & 1174\std{~~5.7} & $1.0\times$ \\
      & Fixed-point iteration   & \textbf{22.0}\%\std{1.2}  & \textbf{32.2}\std{1.7} & \best{4.5}$\times$ &\textbf{28.0}\%\std{1.8} &  \textbf{327}\std{19.9} & \best{3.6}$\times$ \\
      & + Forecasting ($T=1$)   & 36.9\%\std{2.7} & 57.3\std{4.2} & $2.5\times$ & 46.5\%\std{1.9} & 547\std{22.4} & $3.8\times$\\
      \midrule                                                                                             
    CIFAR10 (5 bit)             
      & Baseline                & 100.0\%\std{0.0} & 148.2\std{0.5} &  $1.0\times$ &100.0\%\std{0.0} & 1114\std{~~3.6} & $1.0\times$ \\
      & Fixed-point iteration   & \textbf{15.6}\%\std{2.1}  & \textbf{23.3}\std{3.0} & \best{6.4}$\times$ & \textbf{16.7}\%\std{0.4} & \textbf{239}\std{~~0.6}      & \best{4.7}$\times$ \\
      & + Forecasting ($T=1$)   & 23.2\%\std{2.9} & 35.6\std{4.3} & $4.2\times$ & 27.5\%\std{1.1} & 311\std{10.4} & $3.6\times$\\
      \midrule                                                             
    CIFAR10 (8 bit)             
      & Baseline                & 100.0\%\std{0.0} & 145.7\std{0.8} & $1.0\times$ & 100.0\%\std{0.0} & 1174\std{~~5.7} & $1.0\times$ \\
      & Fixed-point iteration   & \textbf{22.0}\%\std{2.0} & \textbf{32.0}\std{2.9} & \best{4.6}$\times$ & \textbf{25.9}\%\std{1.1} & \textbf{305}\std{11.4} & \best{3.8}$\times$\\ 
      & + Forecasting ($T=1$)   & 43.1\%\std{5.5}  & 65.1\std{8.2} & $2.2\times$ & 50.9\%\std{1.8} & 597\std{21.6} & $2.0\times$\\
      & + Forecasting ($T=5$)   & 59.8\%\std{2.9} & 94.5\std{4.4} & $1.5\times$ & 67.2\%\std{0.6} & 842\std{~~6.2} & $1.4\times$ \\
    \bottomrule                        
    \end{tabular}
    }
\end{table*}

Predictive sampling is evaluated in two settings. 
First, an ARM is trained on images, we refer to this task as explicit likelihood modeling. 
Second, an ARM is trained on the discrete latent space of on autoencoder.

The used datasets are Binary MNIST \cite{larochelle2011neural}, SVHN \cite{netzer2011reading}, CIFAR10 \cite{cifar10}, and ImageNet32 \cite{oord2016pixel}. 
We use the standard test split as test data, except for Imagenet32, for which no test split is available and we use the validation split as test data.
As validation data, we use the last 5000 images of the train split for MNIST and CIFAR10, we randomly select 8527 images from the train split for SVHN, and we randomly select $20000$ images from the train split for Imagenet32.
For all datasets, the remainder of the train split is used as training data. 

The ARM architecture is based on \cite{salimans2017pixelcnn}, with the fully autoregressive categorical output distribution of \cite{oord2016pixel}. 
The categorical output distribution allows us to scale to an arbitrary number of channels without substantial changes to the network architecture.
The autoregressive order is a raster-scan order, and in each spatial location an output channel is dependent on all preceding input channels.

All experiments were performed using PyTorch version 1.1.0 \cite{patszke2019pytorch}.
Training took place on Nvidia Tesla V100 GPUs.
To obtain sampling times, measurements were taken on a single Nvidia GTX 1080Ti GPU, with Nvidia driver 410.104, CUDA 10.0, and cuDNN v7.5.1.
\aw{Sampling results are obtained without caching \cite{ramachandran2017fastgeneration}.}
For a full list of hyperparameters and data preprocessing steps, see Appendix~\ref{app:architecture_and_hyperparams}.


\subsection{Predictive sampling of image data}
\label{sec:predictivesampling:imagedata}

\paragraph{Setting} In this section the performance of predictive sampling for explicit likelihood modelling tasks is tested on binary MNIST, SVHN and CIFAR10. We use the same architecture for all datasets but binary MNIST, for which we decrease the number of layers and filters to prevent overfitting. 
Each ARM is optimized using the log-likelihood objective and performance is reported in bits per dimension (bpd), which is the negative log-likelihood in base two divided by the number of dimensions. 
After 200000 training iterations, the test set performance of the ARMs is $0.150$ bpd on binary MNIST, $1.78$ bpd on SVHN, $1.38$ bpd on CIFAR10 5-bit and $3.08$ bpd on CIFAR10 8-bit. 
Further details on the architecture and optimization procedure are described in Appendix~\ref{app:architecture_and_hyperparams}.

\begin{figure}[t!]
    \centering
    \begin{subfigure}{0.48\textwidth}
        \centering
        \includegraphics[width=\textwidth]{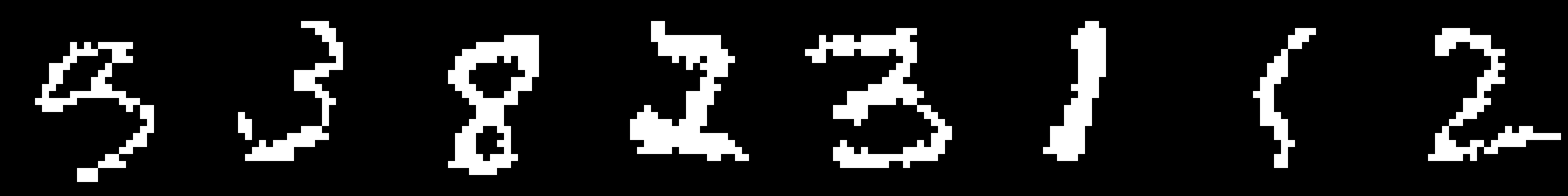}
        \caption{Samples from the model distribution $\rvx \sim \pARM(\,\cdot\,)$.}
        \label{fig:bmnist_explicit_samples_a}
    \end{subfigure}
    \begin{subfigure}{0.48\textwidth}
        \centering
        \includegraphics[width=\textwidth]{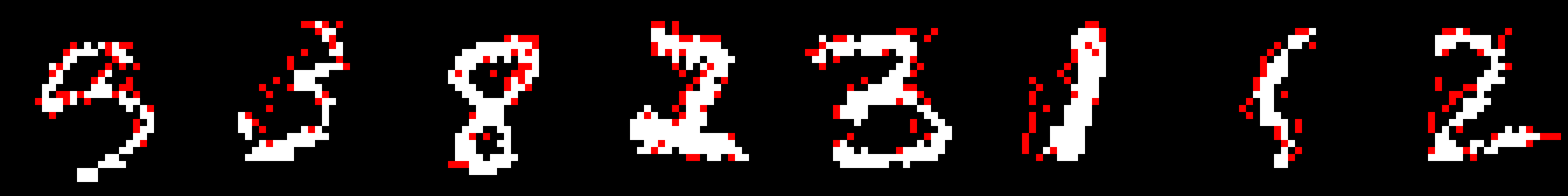}
        \caption{Forecasting mistakes by the forecasting modules.}
        \label{fig:bmnist_explicit_samples_b}
    \end{subfigure}
    \begin{subfigure}{0.48\textwidth}
        \centering
        \includegraphics[width=\textwidth]{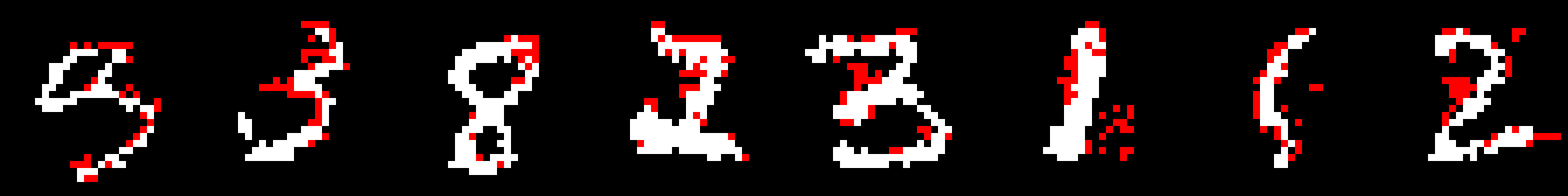}
        \caption{Forecasting mistakes by fixed-point iteration.}
        \label{fig:bmnist_explicit_samples_c}
    \end{subfigure}
    \caption{Samples from the 1-bit ARM. Forecasting mistakes are shown in red.}
    \vspace{-3mm}
    \label{fig:bmnist_explicit_samples}
\end{figure}

For the forecasting modules, we choose a lightweight network architecture that forecasts $T$ future timesteps. 
A triangular convolution is applied to $\rvh$, the hidden representation of the ARM. This is followed by a $1\times1$ convolution with a number of output channels equal to the number of timesteps to forecast multiplied by the number of input categories.
The number of forecasting modules $T$ is $20$ for binary MNIST and $1$ or $5$ for other datasets (the exact number is specified in brackets in the results). 
Forecasts for all remaining future timesteps are taken from the ARM output, as this does not require additional computation.

\begin{figure}[t!]
    \centering
    \begin{subfigure}{0.48\textwidth}
        \centering
        \includegraphics[width=\textwidth]{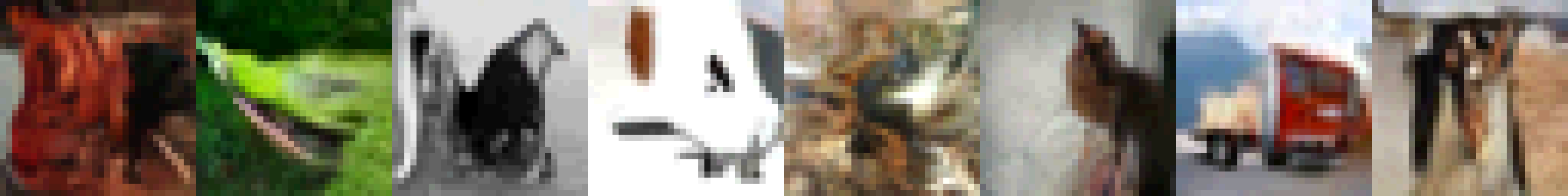}
        \caption{Samples from the model distribution $\rvx \sim \pARM(\,\cdot\,)$.}
        \label{fig:cifar_explicit_samples_a}
    \end{subfigure}
    \begin{subfigure}{0.48\textwidth}
        \centering
        \includegraphics[width=\textwidth]{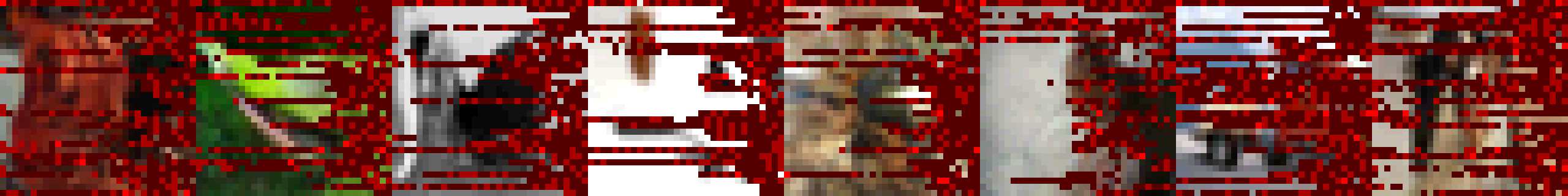}
        \caption{Forecasting mistakes by the forecasting modules.}
        \label{fig:cifar_explicit_samples_b}
    \end{subfigure}
    \begin{subfigure}{0.48\textwidth}
        \centering
        \includegraphics[width=\textwidth]{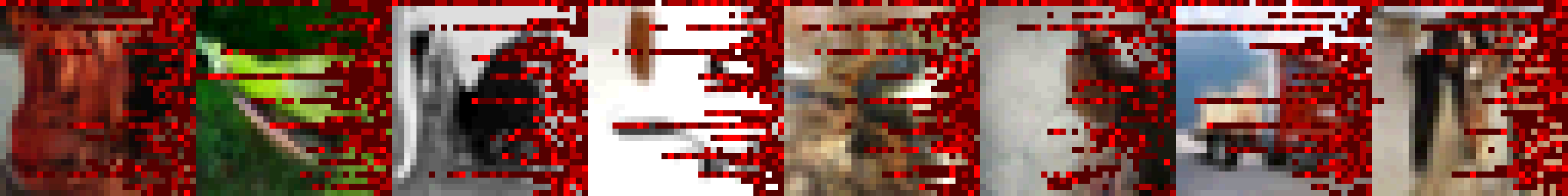}
        \caption{Forecasting mistakes by fixed-point iteration.}
        \label{fig:cifar_explicit_samples_c}
    \end{subfigure}
    \caption{Samples from the 5-bit ARM. The shade of red indicates the number of forecasting mistakes for that location.}
    \vspace{-3mm}
    \label{fig:cifar_explicit_samples}
\end{figure}

\paragraph{Performance}
Sampling performance for ARMs is presented in Table \ref{tab:performance_explicit}. 
For each dataset, we list the percentage of ARM calls with respect to the default sampling procedure, as well as the total runtime during sampling of a batch. 
Results are reported for batch sizes 1 and 32. 
In this implementation, the slowest image determines the number of ARM inference passes.
We leave the implementation of a scheduling system to future work, which would allow sampling at an average rate equal to the batch size 1 setting. 

Fixed-point iteration and learned forecasting greatly outperform the standard baseline on all datasets. 
To put the improvements in perspective, we introduce two additional baselines for binary MNIST: \emph{forecast zeros} and \emph{predict last}. 
The first baseline simply forecasts $\tilde{x}_{i+t} = 0$ for all future timesteps $t$, and the second baseline repeats the last observed value $\tilde{x}_{i+t} = x_{i-1}$. 
On binary MNIST, both fixed-point iteration and learned forecasting outperform these baselines. 

\begin{table*}[t!]
    \centering
    \caption{Performance of predictive sampling for ARMs trained on the latent space of an autoencoder, in terms of percentage of forward passes with respect to the original sampling procedure, and total time to sample. 
    All reported times are based on own implementation.
    Reported means and (Bessel-corrected) standard deviations are based on sampling of 10 batches with random seeds $\{ 0, \ldots, 9 \}$.
    }
    \label{tab:performance_vae}
    \scalebox{0.875}{
    \begin{tabular}{p{2cm} p{3.5cm} r r r r r r r }
\toprule
      &                        & \multicolumn{3}{c}{Batch size 1}     & \multicolumn{3}{c}{Batch size 32} \\
      &                        & ARM calls    & Time (s)  & Speedup   & ARM calls  & Time (s)   & Speedup \\
  \midrule 
    SVHN                 
      & Baseline               & 100.0\%\std{0.0} & 12.1\std{0.0} & 1.0$\times$ & 100.0\%\std{0.0} & 12.6\std{0.2} & 1.0$\times$ \\
      & Fixed-point iteration   & \textbf{15.0}\%\std{2.6} &\textbf{1.9}\std{0.3} & \best{6.4}$\times$ & \textbf{20.3}\%\std{1.2} & \textbf{3.1}\std{0.2} & \best{4.1}$\times$ \\
      & + Forecasting ($T=1$)   & 16.9\%\std{2.8} & 2.2\std{0.3} & 5.5$\times$ & 24.9\%\std{2.9} & 3.8\std{0.4} & 3.3$\times$ \\
  \midrule                                      
    CIFAR10 
      & Baseline                & 100.0\%\std{0.0} & 12.1\std{0.0} & 1.0$\times$ & 100.0\%\std{0.0} & 12.7\std{0.1} & 1.0$\times$ \\
      & Fixed-point iteration   & \textbf{17.6}\%\std{2.9} & \textbf{2.2}\std{0.4} & \best{5.5}$\times$ & \textbf{24.3}\%\std{2.0} & \textbf{3.6}\std{0.3} & \best{3.6}$\times$ \\
      & + Forecasting ($T=1$)   & 19.7\%\std{3.7} & 2.6\std{0.4} & 4.6$\times$ & 26.4\%\std{1.6} & 4.0\std{0.2} & 3.2$\times$ \\
  \midrule
    ImageNet32 
      & Baseline                & 100.0\%\std{0.0} & 12.1\std{0.0} & 1.0$\times$ & 100.0\%\std{0.0} & 12.9\std{0.0} & 1.0$\times$ \\
      & Fixed-point iteration   & \textbf{13.8}\%\std{3.1} & \textbf{1.8}\std{0.3} & \best{6.7}$\times$ & \textbf{20.9}\%\std{2.6} & \textbf{3.1}\std{0.3} & \best{4.2}$\times$ \\
      & + Forecasting ($T=1$)   & 14.2\%\std{2.0} & 1.9\std{0.4} & 6.4$\times$ & 23.0\%\std{2.3} & 3.5\std{0.4} & 3.7$\times$ \\
    \bottomrule                        
    \end{tabular}
    }
\end{table*}

Comparing the sampling speed for 5-bit and 8-bit CIFAR, we observe that when data has a lower-bit depth, it is generally easier to predict future variables.
This can likely be attributed to the lower number of categories.
Typically SVHN is considered to be an easier dataset to model than CIFAR10, a claim which is also supported by the \aw{bpd} of 1.81 for SVHN versus 3.05 for CIFAR10. 
Interestingly, we find that SVHN is not necessarily easier in the case of predictive sampling. 
Comparing the ARM calls for SVHN and CIFAR10 when using fixed-point iteration, both models require approximately 22\% of the ARM calls. 
This suggests that the performance of predictive sampling depends mostly on the number of categories and less on the modeling difficulty of the data. 
Furthermore, while forecasting seems to work well for binary MNIST, the results do not transfer to the more complicated datasets.
For CIFAR10, we observe that increasing the number of forecasting modules decreases performance.
Note also that for binary MNIST the runtime overhead of the forecasting modules negates the effect of the reduced number of ARM inference.

\aw{Note that results are obtained \emph{without} caching \cite{ramachandran2017fastgeneration}.
Combining predictive sampling with caching has the potential to further reduce sampling time.
For example, \cite{ramachandran2017fastgeneration} report a speedup of $2\times$ for batch size 1 and $33\times$ for batch size 32 for PixelCNN++ models trained on CIFAR10.}

To aid quantitative analysis, model samples and corresponding forecasting mistakes are visualized in Figure~\ref{fig:bmnist_explicit_samples} and \ref{fig:cifar_explicit_samples}. 
In these figures, red pixels highlight in which locations in the image the forecast was incorrect, for both forecasting modules and fixed-point iteration. 
As color images consist of three channels, mistakes are visualized using $\frac{1}{3}$, $\frac{2}{3}$ or $\frac{3}{3}$ red, depending on the number of channels that were predicted incorrectly.  
\aw{For binary MNIST samples (Figure~\ref{fig:bmnist_explicit_samples}) it is noticeable that forecasting mistakes do not only lie on the edge of the digits, and that transitions from digit to background pixel are often predicted correctly.
This indicates that more sophisticated patterns are used than, for example, simply repeating the last observed value.}
For more complicated 5-bit CIFAR data (Figure~\ref{fig:cifar_explicit_samples}) we observe more mistakes in the top row and on the right side of the images. 
An explanation for this may be that the ARM dependency structure is from left to right, and top to bottom. 
The left-most pixels are conditioned more strongly on pixels directly above, and these are generally further away in the sequence. 
Hence, even if the last pixel of the preceding row contains a wrong value, pixels in the left-most column can be predicted with high accuracy.


\subsection{Predictive sampling of latent variables}

\paragraph{Setting} In this section we explore autoencoders with a probabilistic latent space \cite{theis2017lossy, balle2017endtoend}. 
Typically these methods weigh a distortion component $\ell$ and a rate component $\log \mathrm{P}(\rvz)$:
\begin{equation}
    \left| \vphantom{\sum} \ell(\rvx, G(\rvz)) - \beta \log \mathrm{P}(\rvz) \right|_{z = Q(\rvx)}
\end{equation}
where $Q$ is an encoder, $G$ is a decoder and $\beta$ is a tunable parameter. 
In our experiments we use the Mean Squared Error (MSE) as distortion metric and set $\beta = 0.1$. 
Following \cite{oord2017neural,habibian2019video,razavi2019generating} we model the latent distribution $\mathrm{P}(\rvz)$ using an ARM. 
The (deterministic) encoder $Q$ has an architecture consisting of two $3 \times 3$ convolutional layers, two strided convolutions and two residual blocks, following PyTorch BasicBlock implementation \cite{he2016deep}. 
The decoder $G$ mirrors this architecture with two residual blocks, two transposed convolutions and two standard convolutional layers. 
The latent space is quantized using an argmax of a softmax, where the gradient is obtained using the straight-through estimator. 
We use a latent space of $4$ channels, with height and width equal to $8$, and $128$ categories per latent variable. 
Further details on the architecture are given in Appendix \ref{app:architecture_and_hyperparams}.  

Following \citet{oord2017neural}, we separate the training of autoencoder and ARM. 
We first train the discrete autoencoder for $50000$ iterations, then freeze its weights, and train an ARM on the latents generated by the encoder for another $150000$ iterations. 
We find that this scheme results in more stability than joint training. 
The obtained MSE is $0.0129$ for Imagenet32, $0.0065$ for CIFAR10, and $0.0008$ for SVHN.
The obtained bits per image dimension are $0.223$ for Imagenet32, $0.244$ for CIFAR10, and $0.191$ for SVHN
(To obtain the bits per latent dimension, multiply these by the dimensionality reduction factor $12$).
Note that the prior likelihood depends on the latent variables produced by the encoder, and cannot be compared directly with results from explicit likelihood modeling.

\paragraph{Performance}
The sampling performance for PixelCNN trained on the latent space of the discrete-latent autoencoder is presented in Table~\ref{tab:performance_vae}. 
\aw{Similar to the explicit likelihood modeling setting}, predictive sampling with fixed-point iteration and learned forecasting modules both outperform the baseline, and fixed-point iteration outperforms learned forecasting across all three datasets.

\begin{figure}[t!]
    \centering
    \begin{subfigure}{0.47\textwidth}
        \centering
        \includegraphics[width=\textwidth]{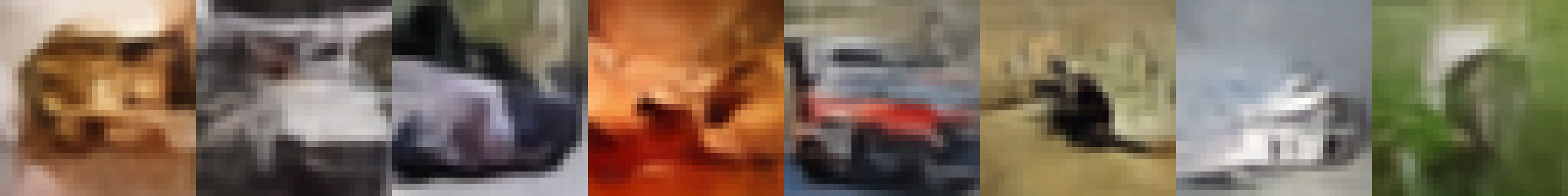}
        \caption{Decoded samples $G(\rvz)$, where $\rvz \sim \mathrm{P}(\rvz)$.}
        \label{fig:samples}
    \end{subfigure}
    
    \begin{subfigure}{0.47\textwidth}
        \centering
        \includegraphics[width=\textwidth]{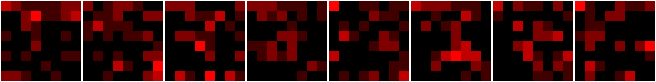}
        \caption{Forecasting mistakes by fixed-point iteration.}
        \label{fig:vae_latent_predsampling_mistakes}
    \end{subfigure}
    
    \begin{subfigure}{0.47\textwidth}
        \centering
        \includegraphics[width=\textwidth]{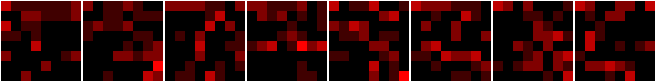}
        \caption{Forecasting mistakes by learned forecasting modules.}
        \label{fig:vae_latent_forecasting_mistakes}
    \end{subfigure}
    \caption{Samples from the VAE and corresponding forecasting mistakes for a $4\times 8 \times 8$ latent space.}
    \vspace{-3mm}
    \label{fig:vae}
\end{figure}

Samples and predictive sampling mistakes of forecasting methods are depicted in Figure \ref{fig:vae} for an autoencoder trained on CIFAR10 (8 bit). 
Samples $\rvz \sim \mathrm{P}(\rvz)$ are generated in the latent representation and subsequently $\hat{\rvx} = G(\rvz)$ is visualized. 
In addition, the latent representation is visualized on a scale from black to red, where the amount of red indicates the number of mistakes at that location, averaged over the channel dimension. 
The latent representation has an $8\times 8$ resolution and is resized to match the $32 \times 32$ images.

\begin{figure}[t!]
    \centering
    \includegraphics[width=0.45\textwidth,trim={0mm 2mm 0mm 0mm}]{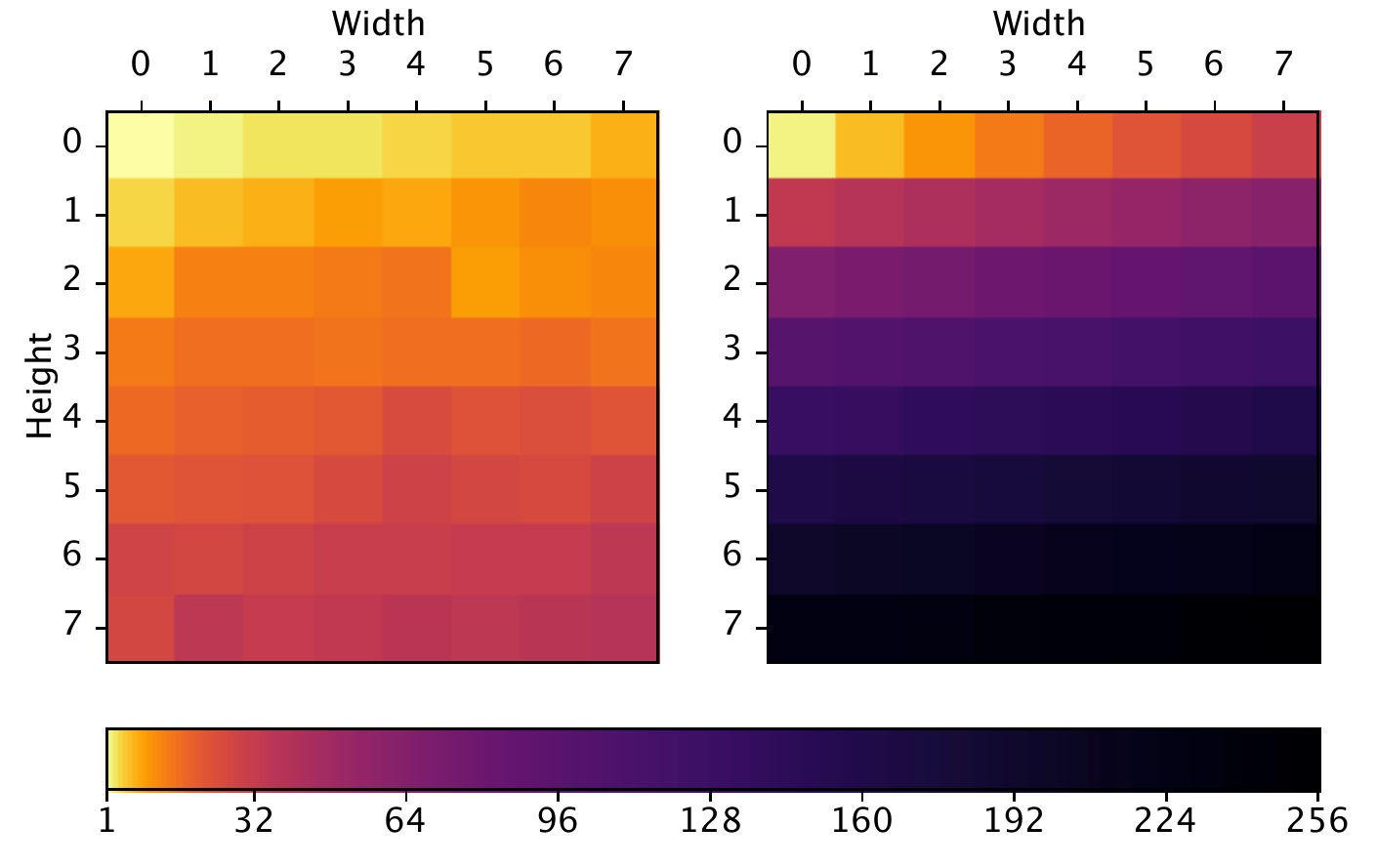}
     \caption{
         Comparison of convergence for fixed-point iteration (left) and the baseline (right) for a $4 \times 8 \times 8$ latent space of an autoencoder trained on CIFAR10.
         Each spatial location shows the iteration at which the final value was determined, averaged over all latent channels and over 32 samples $\rvz \sim \mathrm{P}(\rvz)$.
         Note that a log-scale colormap is used to emphasize differences for low values.
    }
    \label{fig:heatmap_convergence}
      \vspace{-1mm}
\end{figure}

Finally, the convergence behavior of fixed-point iteration is visualized in Figure \ref{fig:heatmap_convergence}.
In this figure, the color indicates the iteration of sampling from which the variable remained the same, \textit{i.e.}, the iteration at which that variable converged. 
For example, because there is strict triangular dependence and the top-left variable in the first channel is at the beginning of the sequence, this variable will converge at step one. 
The converging iterations are averaged over channels and a batch of $32$ images. 
The right image of Figure \ref{fig:heatmap_convergence} shows the baseline, where the total number of iterations is exactly equal to the number of dimensions. 
The left image shows the convergence of the ARM fixed-point iteration procedure, which needs $53$ iterations on average for this batch of data.
We observe that pixels on the left of the image tend to converge earlier than those on the right. 
This matches the conditioning structure of the ARM, where values in the left-most column depend strongly on pixel values directly above, and right-most variables also depend on pixels to their left.

\begin{table}[t!]
    \centering
    \caption{Ablation showing the effect of reparametrization and representation sharing for CIFAR10. Means and (Bessel-corrected) standard deviations are based on 10 sampled batches of size 32, with random seeds $\{0, \ldots 9\}$.}
    \label{tab:ablations}
  \scalebox{.875}{
    \begin{tabular}{ l r r r r }
    \toprule
         & \multicolumn{2}{c}{CIFAR10} \\ 
         & ARM calls  & Time (s) \\ \midrule
        Fixed-point iteration              & 25.9\%\std{1.1} & 305\std{11.4} \\ 
         ~ without reparametrization       & 97.2\%\std{0.4} & 1122\std{~~6.4} \\ \midrule
        Learned forecasting                & 50.9\%\std{1.8} & 597\std{21.6} \\
         ~ without representation sharing  & 67.1\%\std{3.3} & 802\std{19.5} \\
    \bottomrule
    \end{tabular}
  }
  \vspace{-3mm}
\end{table}

\subsection{Ablations}
We perform ablations on 8 bit CIFAR10 data to show the effect of the isolation of stochasticity via reparametrization, and the sharing of the ARM representation. 
First, to quantify the effect of reparametrization, the sampling procedure is run again for an ARM without learned forecasting modules.
As forecast, the most likely value according the forecasting distribution $\pFARM$ is used. 
For categorical distributions, this is done by removing the $\epsilon_{i,c}$ term from Equation \ref{eq:forecast_arm_categorical}. 
In addition, we show the importance of sharing the ARM representation by training forecasting modules conditioned only on $\rvx_{<i}$ and reparametrization noise $\boldsymbol{\epsilon}$, \textit{i.e.}, Equation \ref{eq:forecasting_sharing} where $\rvh$ is removed. 
Results are shown in Table~\ref{tab:ablations}. 
These experiments indicate that the both reparametrization and the shared representation improve performance considerably, with reparametrization having the biggest effect.

\section{Conclusion}

We introduce predictive sampling, an algorithm speeds up sampling for autoregressive models (ARMs), while keeping the model intact.
The algorithm aims to forecast likely future values and exploits the parallel inference property of neural network based ARMs. 
We propose two variations to obtain forecasts, namely ARM fixed-point iteration and learned forecasting modules.
In both cases, the sampling procedure is reduced to a deterministic function by a reparametrization.
We train ARMs on image data and on the latent space of a discrete autoencoder, and show in both settings that predictive sampling provides a considerable increase in sampling speed. 
ARM fixed-point iteration, a method that requires no training, obtains the best performance overall.

\bibliography{references}
\bibliographystyle{icml2020}

\clearpage
\appendix
\section{Architecture and hyperparameters}
\label{app:architecture_and_hyperparams}

\subsection{Autoregressive model architecture}

We base our PixelCNN implementation on a Pytorch implementation of PixelCNN++ \cite{salimans2017pixelcnn} by GitHub user pclucas14 (\href{https://github.com/pclucas14/pixel-cnn-pp}{https://github.com/pclucas14/pixel-cnn-pp, commit 16c8b2f}).
We make the following modifications.

Instead of the discretized mixture of logistics loss as described in \citep{salimans2017pixelcnn}, we utilize the categorical distributions as described in \citep{oord2016pixel}, which allows us to model distributions with full autoregressive dependence. This is particularly useful when training the PixelCNN for discrete-latent autoencoders, as the number of input channels can be altered without substantial changes to the implementation.  
We model dependencies between channels as in the PixelCNN architecture, by masking convolutions so that the causal structure is preserved.
That is, the output corresponding to input $x_{c,h,w}$ is conditioned on all previous rows ${x_{0,:,:}, \ldots, x_{h-1,:,:}}$, on all previous columns of the same row ${x_{h,0,:}, \ldots, x_{h,w-1,:}}$ and all previous channels of the same spatial location $x_{h,w,0}, \ldots, x_{h,w,c-1}$.

The original implementation normalizes the input data to a range between $-1$ and $1$. 
Instead, we follow \cite{oord2016pixel} and use a one-hot encoding for inputs. 
Additionally, we do not use weight normalization.

\subsection{Forecasting module architecture}

The forecasting module used in this work consists of a single strictly triangular $3\times3$ convolution followed by a $1\times1$ convolution, where the number of output channels is equal to the number of data channels multiplied by the number of categories.
The masked convolution is applied to the last activation of the \emph{up-left stack} of the PixelCNN, $\rvh$.
We set the number of channels for this layer to 162 for image space experiments, and 160 for latent space experiments.

We experimented with variations of forecasting modules that use $x$ (one-hot) or truncated Gumbel noise, obtained as described in Section \ref{app:posterior_reparametrization}, as additional inputs.
For the forecasting module capacity we considered, this did not lead to improved sampling performance.

\subsection{Default hyperparameters}

\paragraph{Explicit likelihood modeling}
Hyperparameter settings for the PixelCNNs trained on image data are given in Table \ref{tab:arm_parameters}. 
We use the same parameters across all datasets to maintain consistency, and did not alter the architecture for likelihood performance.
The exception is binary MNIST, where we observed strong overfit if the size was not changed.

\begin{table}[t!]
    \centering
    \caption{Hyperparameters for the trained PixelCNN models.}
    \label{tab:arm_parameters}
    \scalebox{0.9}{
        \centering 
        \begin{tabular}{ l r r r }
        \toprule
             Hyperparameter            & Binary MNIST & Default        \\
            \hline 
             Learning rate             & 0.0002 & 0.0002         \\
             Learning rate decay       & 0.999995 & 0.999995       \\
             Batch size                & 64       & 64             \\
             Max iterations            & 200000   & 200000         \\
             Weight decay              & 1e-6     & 1e-6           \\
             Optimizer                 & Adam     & Adam           \\
            \hline                               
             Number of gated resnets   &  2     & 5              \\
             Filters per layer         & 60     & 162            \\
             Dropout rate              & 0.5    & 0.5            \\
             Nonlinearity              & concat\_elu & concat\_elu    \\
            \hline                             
             Forecasting modules       &  20    & 1              \\
             Forecasting filters       &  60    & 162            \\ 
             Forecasting loss weight   &  0.01  & 0.01           \\
         \bottomrule
        \end{tabular}
    }
\vspace{-1mm}
\end{table}

\paragraph{Latent space modeling}

For the latent space experiments, we use an encoder and decoder with bottleneck structure, and a PixelCNN to model the resulting latent space.
The \emph{width} of the encoder and decoder, \textit{i.e.}, the parameter that controls the number of channels at every layer, is 512 for all experiments.
The used loss function is Mean Squared Error, and the input data is normalized to the range $[-1, 1]$. 

The encoder consists of the following layers.
First, two $3 \times 3$ convolutional layers with padding 1 and half width.
Then, one strided $4 \times 4$ convolution of half width with padding 1 and stride 2, followed by a similar layer of full width. 
We then apply two residual blocks (PyTorch BasicBlock implementation \cite{he2016deep}).
Finally, a $1\times1$ convolution layer maps to the desired number of latent channels.

The decoder architecture mirrors the encoder architecture.
First, a $1\times1$ convolution layer maps from the (one-hot) latents to the desired width.
Two residual blocks are applied, followed by a full width transpose convolution and a half width transpose convolution, both having the same parameters as their counterparts in the encoder.
Lastly, two $3\times3$ convolution layers of half width are applied, where the last layer has three output channels.

The latent space is quantized by taking the argmax over a softmax, and one-hot encoding the resulting latent variable.
As quantization is non-differentiable, the gradient is obtained using a straight-through estimator, \textit{i.e.}, the softmax gradient is used in the backward pass. 
We use a latent space of $4$ channels, with height and width equal to $8$, and $128$ categories per latent variable. 

Optimization parameters and the parameters of the PixelCNN that is used to model the latent space are kept the same as in the explicit likelihood setting, see Table~\ref{tab:arm_parameters}.
We do not train the autoencoder and ARM jointly. Instead, we train an autoencoder for 50000 iterations, then 
freeze the autoencoder weights and train an ARM on the latent space for an additional 200000 iterations.

\subsection{Infrastructure}
Software used includes Pytorch \cite{patszke2019pytorch} version 1.1.0, CUDA 10.0, cuDNN 7.5.1.
All sampling time measurements were obtained on a single Nvidia 1080Ti GPU using CUDA events, and we only compute runtime after calling \textit{torch.cuda.synchronize}.
Training was performed on Nvidia TeslaV100 GPUs, with the same software stack as the evaluation system.

\section{Posterior Reparametrization Noise}
\label{app:posterior_reparametrization}

To condition forecasting modules on reparametrization noise when training on the data distribution, sample noise pairs ($\rvx, \boldsymbol{\epsilon}$) are needed. 
In principle, these can be created by sampling $\boldsymbol{\epsilon}$ and computing the corresponding $\rvx$ using the ARM, \emph{i.e.}, by computing the autoregressive inverse.
However, this process may be slow, and does not allow for joint training of the ARM and forecasting module. 
Alternatively, one can use the assumption that the model distribution $\mathrm{P}_{\mathrm{ARM}}(\rvx)$ will sufficiently approximate $\mathrm{P}_{\mathrm{data}}(\rvx)$. 
In this case, ($\rvx, \boldsymbol{\epsilon}$) pairs can be sampled using the data distribution $\mathrm{P}_{\mathrm{data}}(\rvx)$, and the posterior of the noise $p(\boldsymbol{\epsilon} | \rvx)$:
\begin{equation}
    \rvx, \boldsymbol{\epsilon} \sim \mathrm{P}_{\mathrm{ARM}}(\rvx | \boldsymbol{\epsilon}) p(\boldsymbol{\epsilon}) \approx p(\boldsymbol{\epsilon} | \rvx) \mathrm{P}_{\mathrm{data}}(\rvx),
\end{equation}
where $\mathrm{P}_{\mathrm{ARM}}(\rvx | \boldsymbol{\epsilon})$ is a Dirac delta peak on the output of the reparametrization, and $p(\boldsymbol{\epsilon} | \rvx)$ denotes the posterior of the noise given a sample $\rvx$:
\begin{equation}
    p(\boldsymbol{\epsilon} | \rvx) = \frac{\mathrm{P}_{\mathrm{ARM}}(\rvx | \boldsymbol{\epsilon}) p(\boldsymbol{\epsilon})}{\mathrm{P}_{\mathrm{ARM}}(\rvx)}.
\end{equation}
In the case of the Gumbel-Max reparametrization, the posterior Gumbel noise $p(\boldsymbol{\epsilon}|\rvx)$ can be computed straightforwardly by using the notion that the maximum and the location of the maximum are independent  \cite{maddison2014sampling,kool2019stochastic}. 
First, we sample from the Gumbel distribution for the arg max locations, \emph{i.e.}, the locations that resulted in the sample $\rvx$:
\begin{equation}
    \epsilon_{i,x_i} \sim G. 
\end{equation}
Subsequently, the remaining values can be sampled using truncated Gumbel distributions ($TG$) \cite{maddison2014sampling,kool2019stochastic}. 
The truncation point is located at the maximum value $\mu_{i,x_i} + \epsilon_{i,x_i}$:
\begin{equation}
    \epsilon_{i,c} \sim TG(\mu_{i,c} | \mu_{i,x_i} + \epsilon_{i,x_i}) - \mu_{i,c} \quad \text{for all} \quad c \not=x_i.
\end{equation}
Here, $\mu_{i,c}$ denotes the logit from the model distribution $\mathrm{P}_{\mathrm{ARM}}(\rvx)$ of dimension $i$ for category $c$. 

To summarize, a sample ($\rvx, \boldsymbol{\epsilon}$) is created by first sampling $\rvx \sim \mathrm{P}_{\mathrm{data}}$ from data, and then sampling $\boldsymbol{\epsilon} \sim p(\boldsymbol{\epsilon} | \rvx)$ using the Gumbel and Truncated Gumbel distributions as described above. 
This technique allows simultaneous training of the ARM and forecasting module conditioned on Gumbel noise without the need to create a dataset of samples.

\section{Generated samples}
\label{app:samples}

We show 16 samples for each of the models trained with forecasting modules, as well as forecasting mistakes.
To find forecasting mistakes made by ARM fixed-point iteration, we simply disable the forecasting modules during sampling.
All samples were generated using the same random seed (10) and were not cherry-picked for perceptual quality or sampling performance.
The used datasets are Binary MNIST \cite{larochelle2011neural}, SVHN \cite{netzer2011reading}, CIFAR10 \cite{cifar10}, and ImageNet32 \cite{oord2016pixel}.

\begin{figure}[t!]
    \centering
    \begin{subfigure}{0.48\textwidth}
        \centering
        \includegraphics[width=\textwidth]{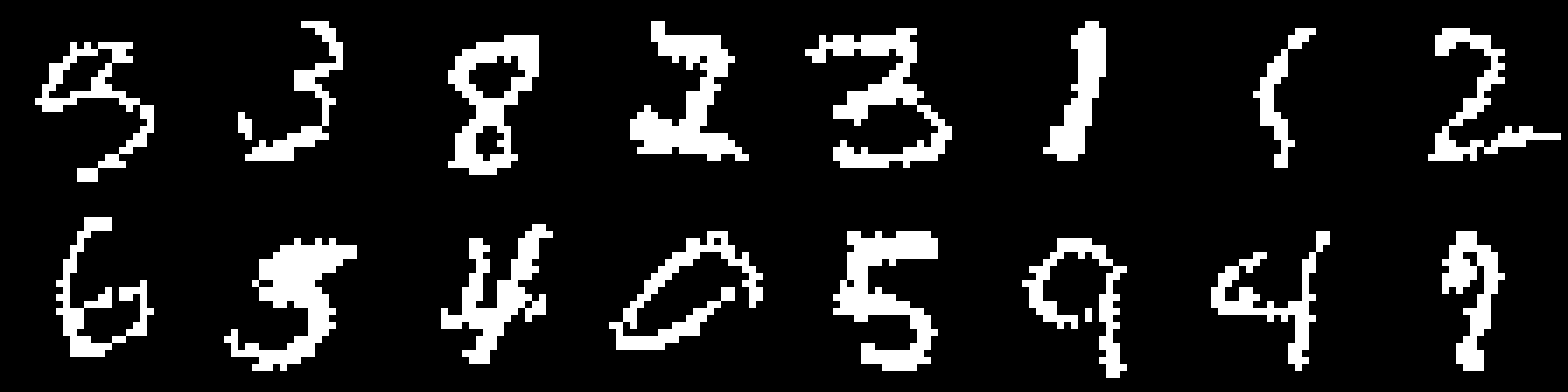}
        \caption{Samples from the model distribution $\rvx \sim \pARM(\,\cdot\,)$.}
    \end{subfigure}
    \begin{subfigure}{0.48\textwidth}
        \centering
        \includegraphics[width=\textwidth]{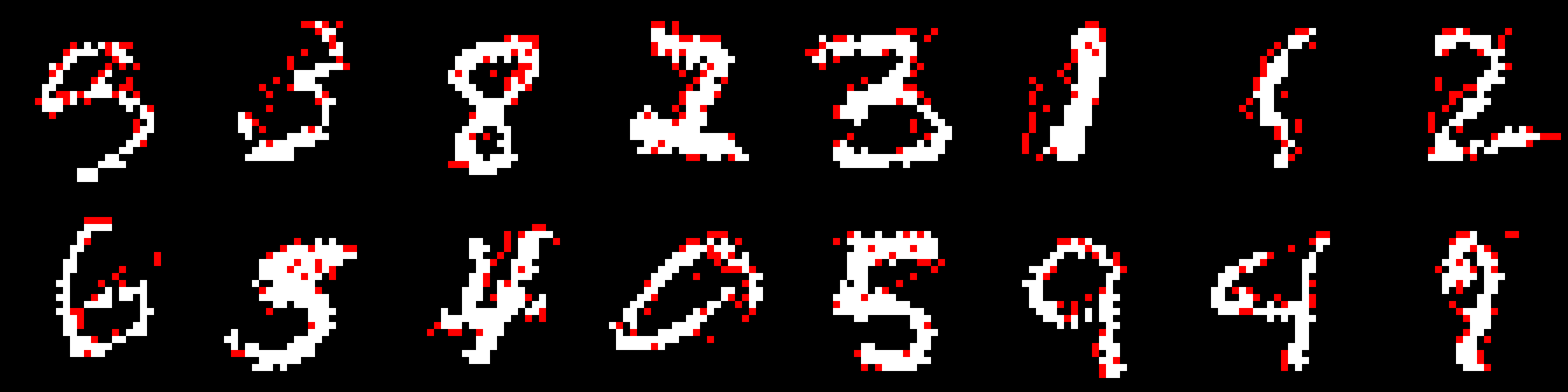}
        \caption{Forecasting mistakes by the forecasting modules.}
    \end{subfigure}
    \begin{subfigure}{0.48\textwidth}
        \centering
        \includegraphics[width=\textwidth]{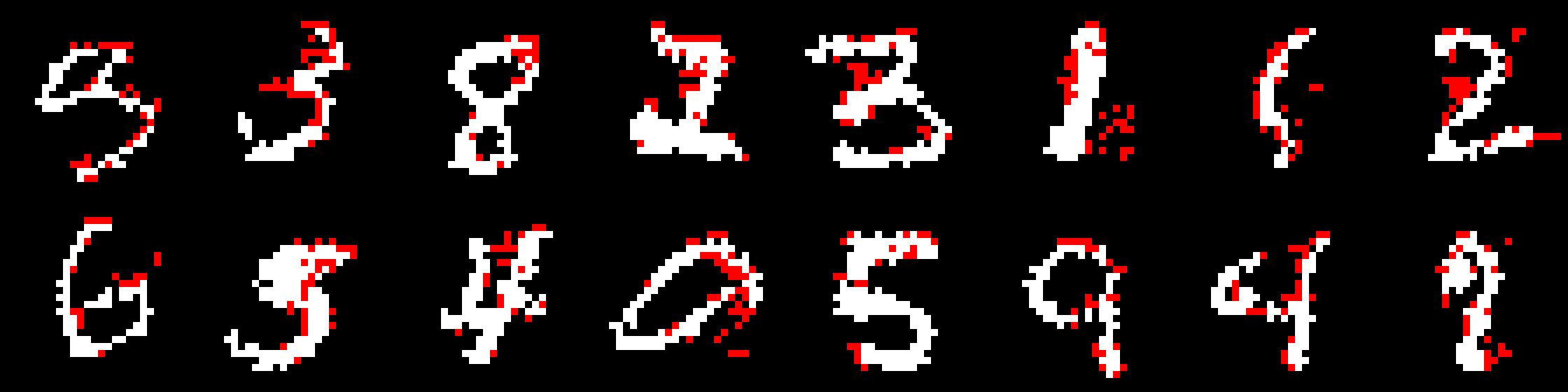}
        \caption{Forecasting mistakes by fixed-point iteration.}
    \end{subfigure}
    \caption{Samples from the 1-bit ARM and forecasting mistakes.}
    \label{app:fig:bmnist_explicit_samples}
\end{figure}

\begin{figure}[t!]
    \centering
    \begin{subfigure}{0.48\textwidth}
        \centering
        \includegraphics[width=\textwidth]{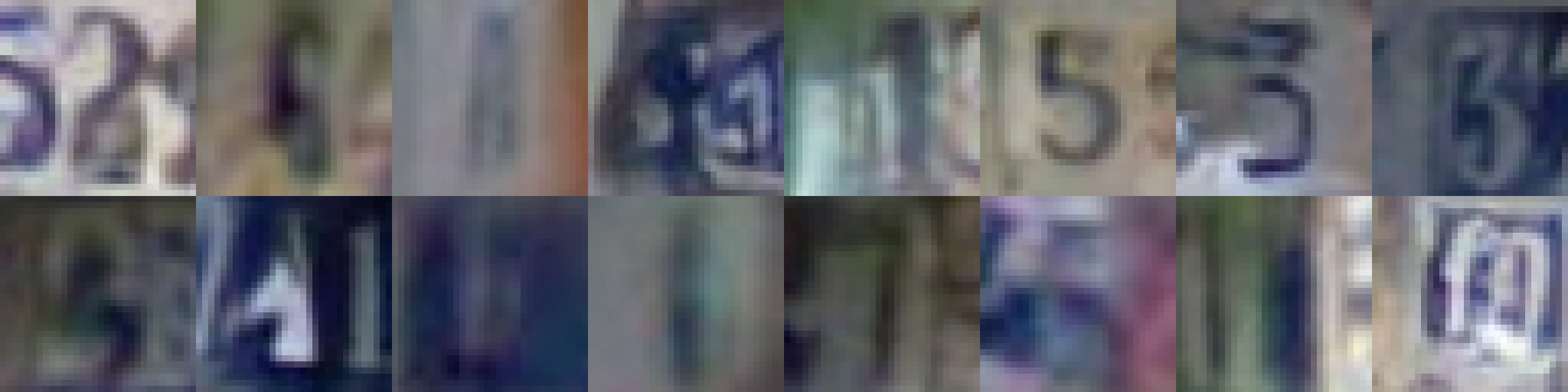}
        \caption{Samples from the model distribution $\rvx \sim \pARM(\,\cdot\,)$.}
    \end{subfigure}
    \begin{subfigure}{0.48\textwidth}
        \centering
        \includegraphics[width=\textwidth]{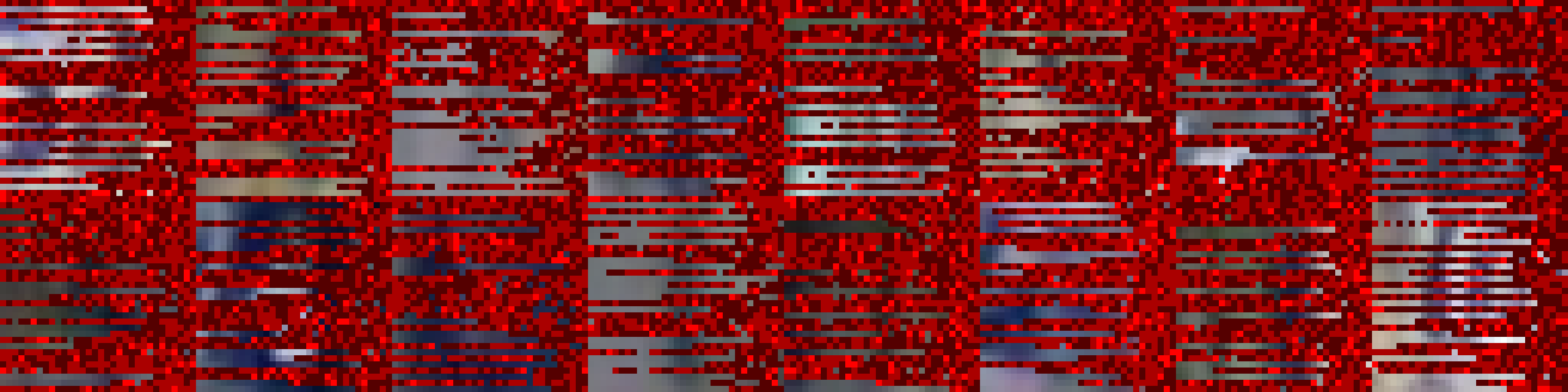}
        \caption{Forecasting mistakes by the forecasting modules.}
    \end{subfigure}
    \begin{subfigure}{0.48\textwidth}
        \centering
        \includegraphics[width=\textwidth]{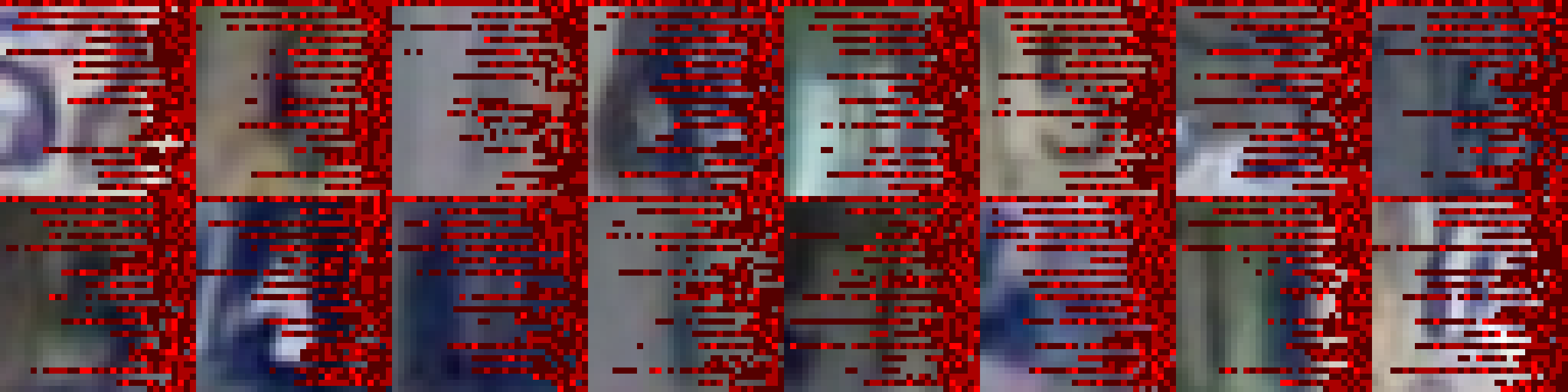}
        \caption{Forecasting mistakes by fixed-point iteration.}
    \end{subfigure}
    \caption{Samples from the 8-bit ARM and forecasting mistakes.}
    \label{app:fig:svhn8_explicit_samples}
\end{figure}

\begin{figure}[t!]
    \centering
    \begin{subfigure}{0.48\textwidth}
        \centering
        \includegraphics[width=\textwidth]{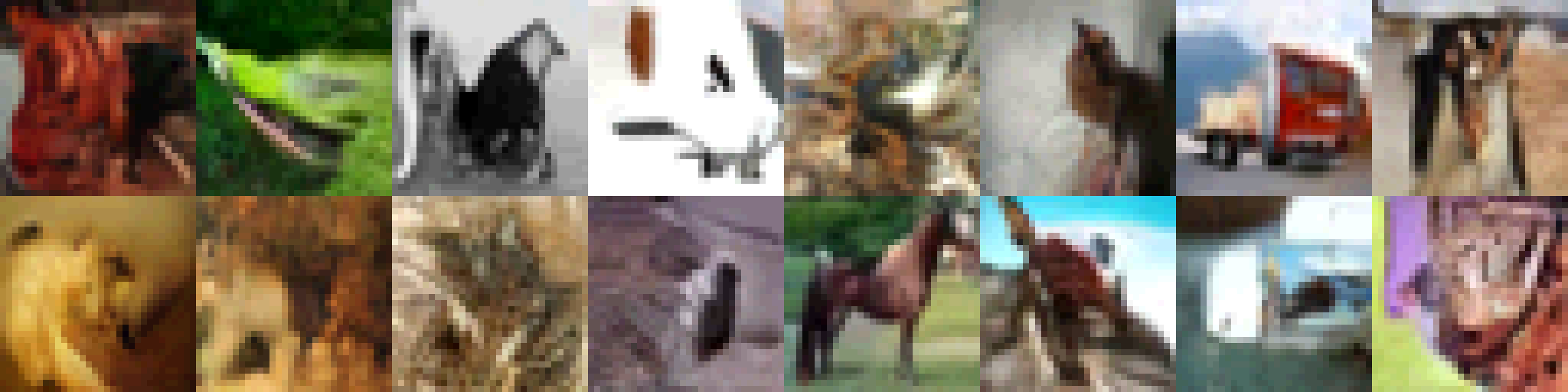}
        \caption{Samples from the model distribution $\rvx \sim \pARM(\,\cdot\,)$.}
    \end{subfigure}
    \begin{subfigure}{0.48\textwidth}
        \centering
        \includegraphics[width=\textwidth]{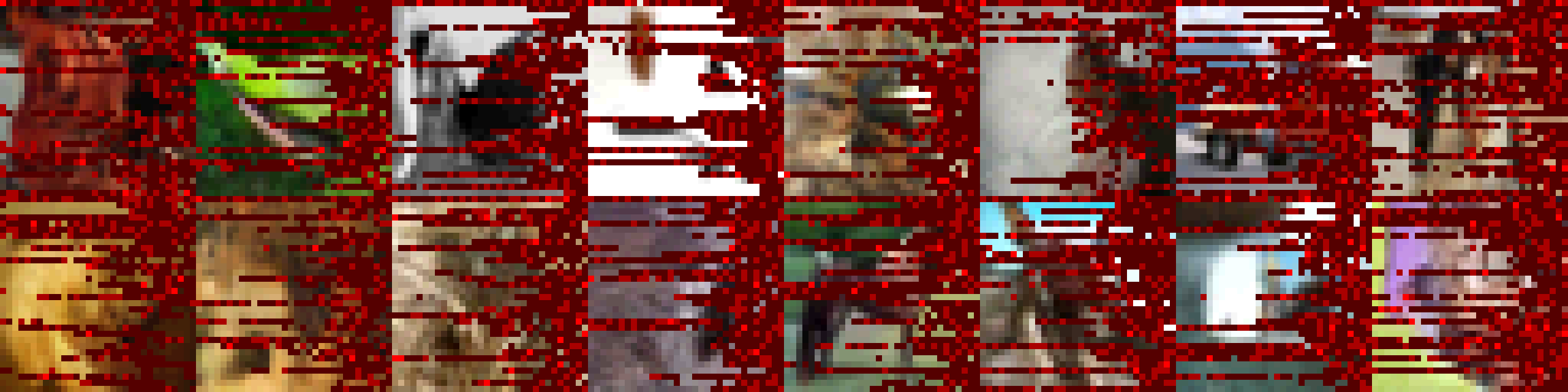}
        \caption{Forecasting mistakes by the forecasting modules.}
    \end{subfigure}
    \begin{subfigure}{0.48\textwidth}
        \centering
        \includegraphics[width=\textwidth]{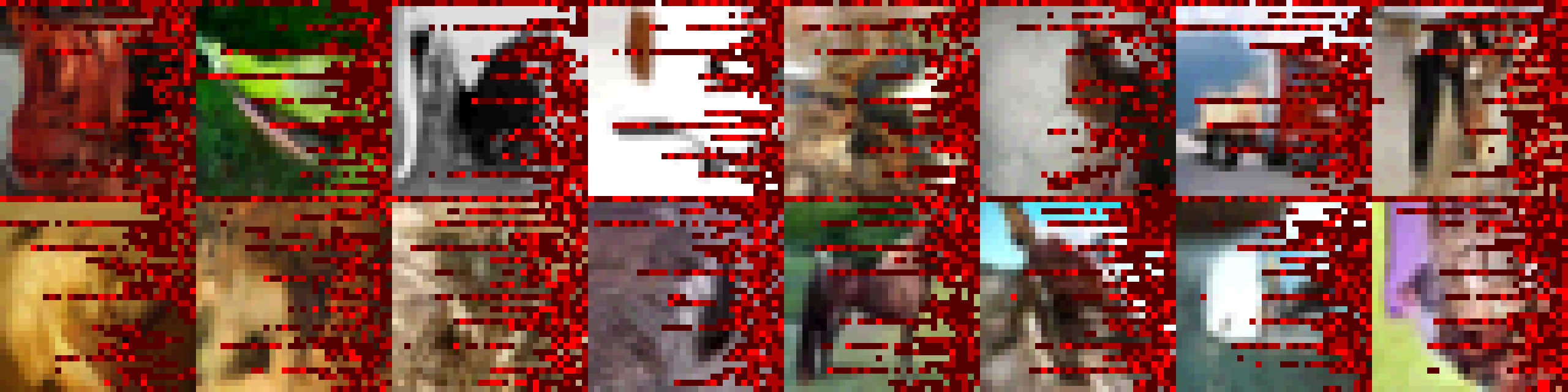}
        \caption{Forecasting mistakes by fixed-point iteration.}
    \end{subfigure}
    \caption{Samples from the 5-bit ARM and forecasting mistakes.}
    \label{app:fig:cifar5_explicit_samples}
\end{figure}

\begin{figure}[ht!]
    \centering
    \begin{subfigure}{0.48\textwidth}
        \centering
        \includegraphics[width=\textwidth]{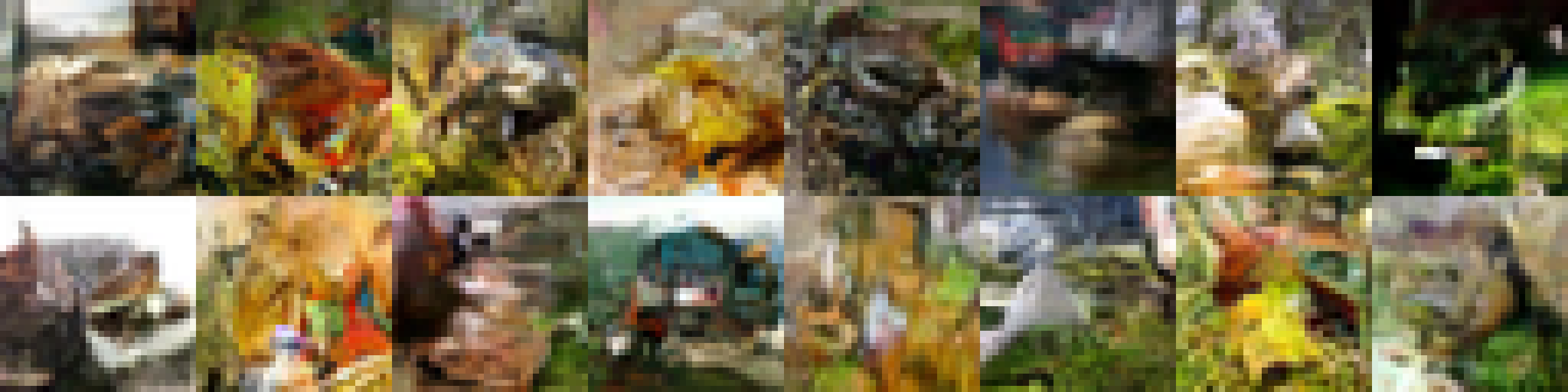}
        \caption{Samples from the model distribution $\rvx \sim \pARM(\,\cdot\,)$.}
    \end{subfigure}
    \begin{subfigure}{0.48\textwidth}
        \centering
        \includegraphics[width=\textwidth]{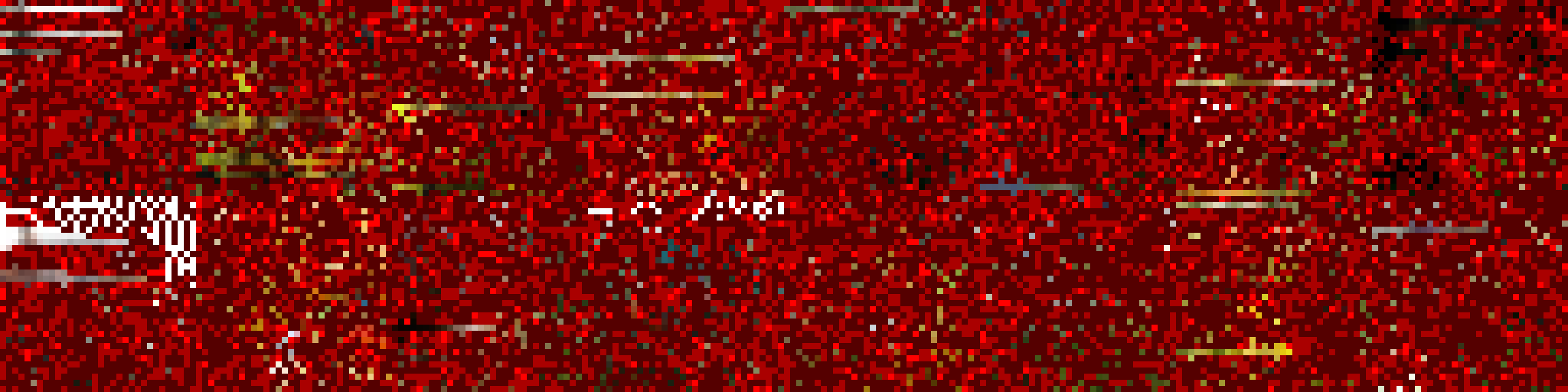}
        \caption{Forecasting mistakes by the forecasting modules.}
    \end{subfigure}
    \begin{subfigure}{0.48\textwidth}
        \centering
        \includegraphics[width=\textwidth]{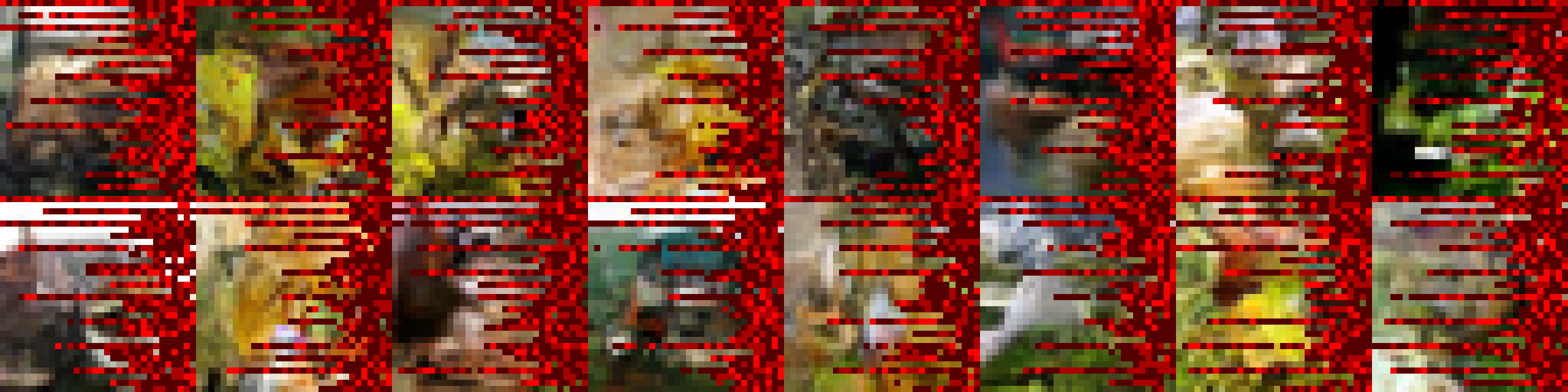}
        \caption{Forecasting mistakes by fixed-point iteration.}
    \end{subfigure}
    \caption{Samples from the 8-bit ARM and forecasting mistakes.}
    \label{app:fig:cifar8_explicit_samples}
\end{figure}

Samples generated by the model trained on binary MNIST are shown in
Figure~\ref{app:fig:bmnist_explicit_samples},
Figure~\ref{app:fig:svhn8_explicit_samples} shows SVHN 8-bit samples, 
and Figures~\ref{app:fig:cifar5_explicit_samples} and \ref{app:fig:cifar8_explicit_samples} show samples generated by the ARM trained on CIFAR10 for 5-bit and 8-bit data, repsectively.

\begin{figure}[t!]
    \centering
    \begin{subfigure}{0.47\textwidth}
        \centering
        \includegraphics[width=\textwidth]{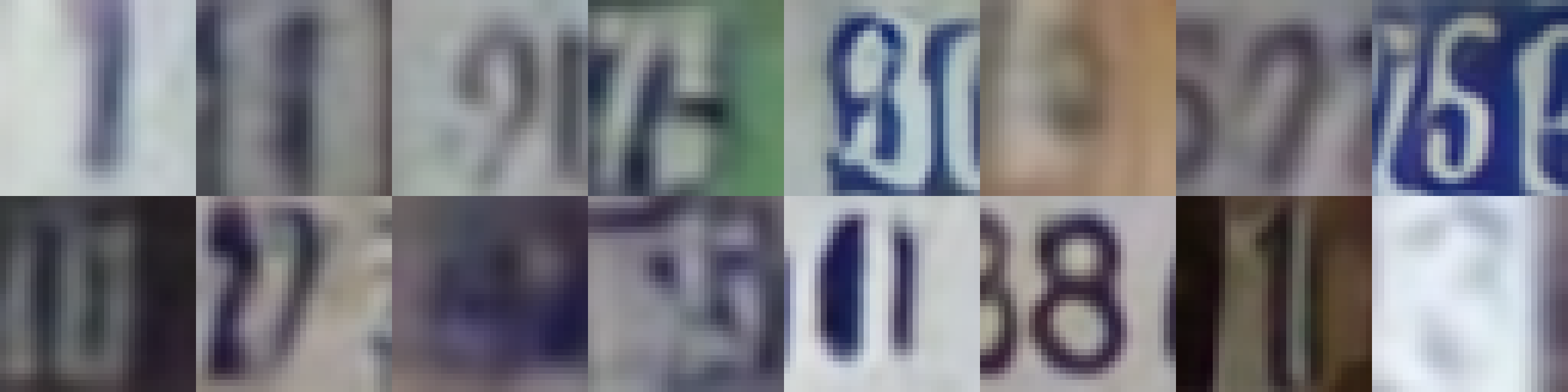}
        \caption{Decoded samples $G(\rvz)$, where $\rvz \sim \mathrm{P}(\rvz)$.}
    \end{subfigure}
    
    \begin{subfigure}{0.47\textwidth}
        \centering
        \includegraphics[width=\textwidth]{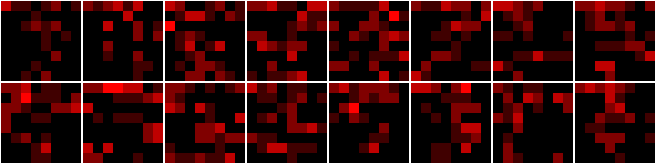}
        \caption{Forecasting mistakes by learned forecasting modules.}
    \end{subfigure}
    
    \begin{subfigure}{0.47\textwidth}
        \centering
        \includegraphics[width=\textwidth]{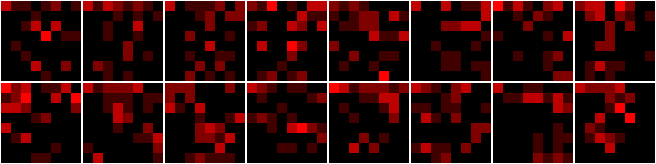}
        \caption{Forecasting mistakes by fixed-point iteration.}
    \end{subfigure}
    
    \caption{VAE samples, and forecasting mistakes in latent space.}
    \label{app:fig:vae_samples_svhn}
\end{figure}

\begin{figure}[t!]
    \centering
    \begin{subfigure}{0.47\textwidth}
        \centering
        \includegraphics[width=\textwidth]{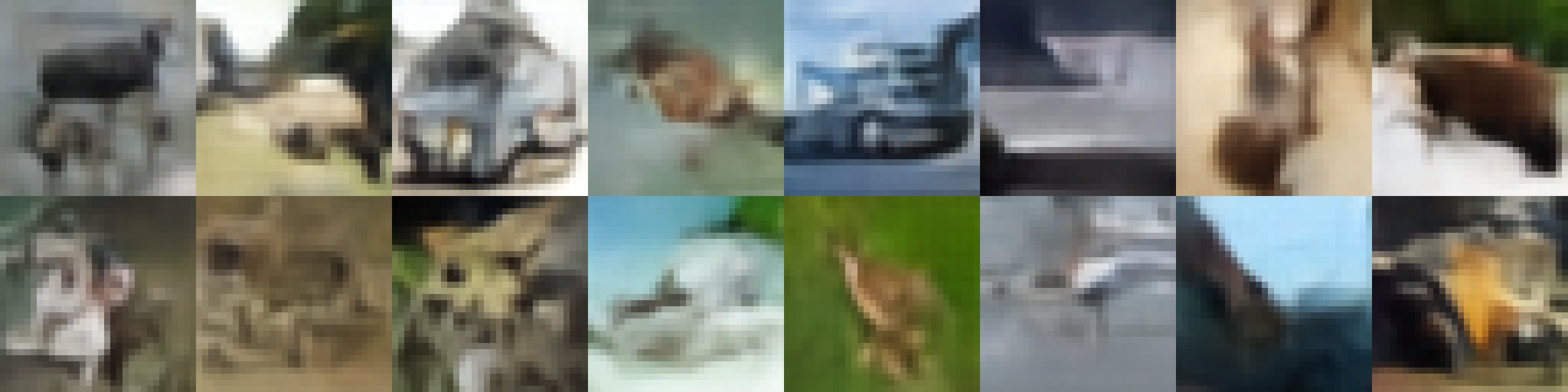}
        \caption{Decoded samples $G(\rvz)$, where $\rvz \sim \mathrm{P}(\rvz)$.}
    \end{subfigure}
    
    \begin{subfigure}{0.47\textwidth}
        \centering
        \includegraphics[width=\textwidth]{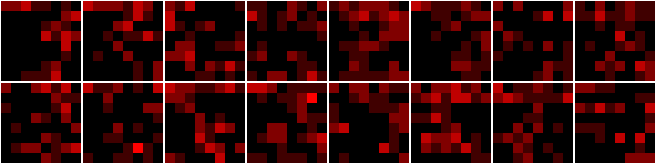}
        \caption{Forecasting mistakes by learned forecasting modules.}
    \end{subfigure}
    
    \begin{subfigure}{0.47\textwidth}
        \centering
        \includegraphics[width=\textwidth]{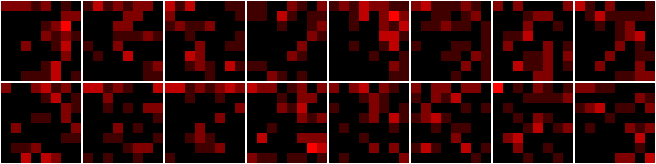}
        \caption{Forecasting mistakes by fixed-point iteration.}
    \end{subfigure}
    
    \caption{VAE samples, and forecasting mistakes in latent space.}
    \label{app:fig:vae_samples_cifar}
\end{figure}

\begin{figure}[t!]
    \centering
    \begin{subfigure}{0.47\textwidth}
        \centering
        \includegraphics[width=\textwidth]{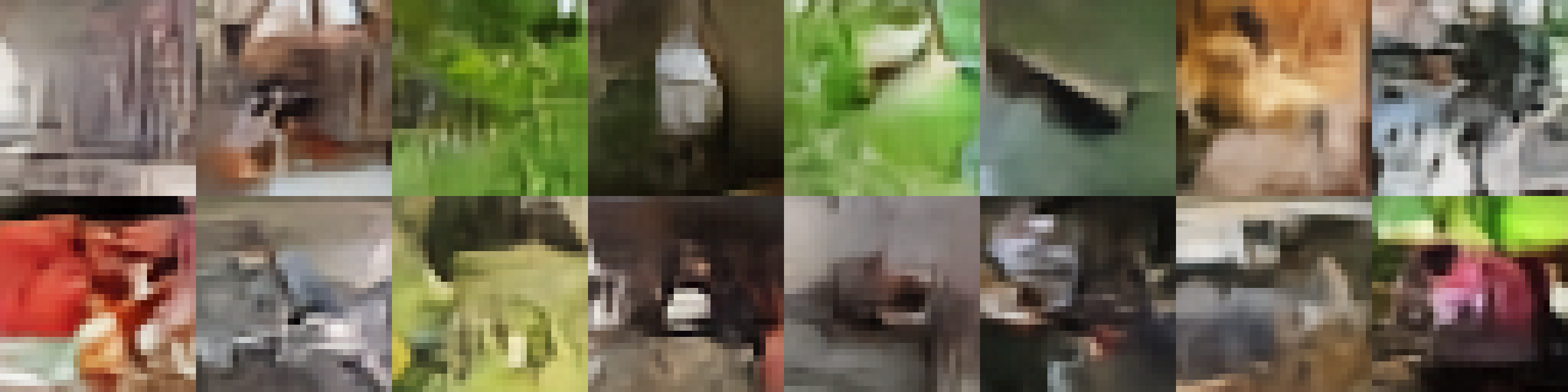}
        \caption{Decoded samples $G(\rvz)$, where $\rvz \sim \mathrm{P}(\rvz)$.}
    \end{subfigure}
    
    \begin{subfigure}{0.47\textwidth}
        \centering
        \includegraphics[width=\textwidth]{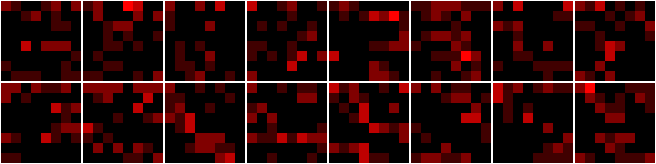}
        \caption{Forecasting mistakes by learned forecasting modules.}
    \end{subfigure}
    
    \begin{subfigure}{0.47\textwidth}
        \centering
        \includegraphics[width=\textwidth]{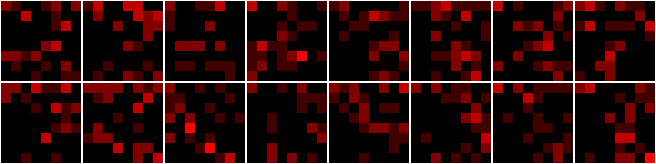}
        \caption{Forecasting mistakes by fixed-point iteration.}
    \end{subfigure}
    
    \caption{VAE samples, and forecasting mistakes in latent space.}
    \label{app:fig:vae_samples_imagenet32}
\end{figure}

Samples from the VAE are generated by decoding a sample from the ARM trained on the latent space.
That is, we first generate a latent variable from the trained ARM $z \sim P(z)$.
This sample is then decoded to image-space using the decoder $G$ as $\hat{x} = G(z)$.
We show samples for SVHN in Figure~\ref{app:fig:vae_samples_svhn}, for CIFAR10 in Figure~\ref{app:fig:vae_samples_cifar}, and for Imagenet32 in Figure~\ref{app:fig:vae_samples_imagenet32}.

\clearpage

\vspace{6mm}

\vspace{6mm}

%
%
%
%
\clearpage

\end{document}